%% file: 3129.tex
\documentclass[runningheads]{llncs}
\usepackage{graphicx}

\usepackage{tikz}
\usepackage{comment}
\usepackage{amsmath,amssymb} 
\usepackage{color}
\usepackage[misc]{ifsym}

\usepackage[accsupp]{axessibility}  

\usepackage[width=122mm,left=12mm,paperwidth=146mm,height=193mm,top=12mm,paperheight=217mm]{geometry}

\usepackage[export]{adjustbox}
\usepackage{subcaption}
\usepackage{amsfonts}
\usepackage{xspace}
\usepackage{multirow}
\usepackage[toc,title,page]{appendix}
\usepackage{graphicx}
\usepackage{floatrow}

\usepackage[colorlinks=true]{hyperref}

\newfloatcommand{capbtabbox}{table}[][\FBwidth]
\usepackage{booktabs,makecell}
\makeatletter
\DeclareRobustCommand\onedot{\futurelet\@let@token\@onedot}
\def\@onedot{\ifx\@let@token.\else.\null\fi\xspace}

\usepackage{pifont}
\usepackage{xcolor}
\definecolor{deemph}{gray}{0.6}

\newcommand{\cmark}{\ding{51}\xspace}%
\newcommand{\xmarkg}{\textcolor{lightgray}{\ding{55}}\xspace}%

\def\etal{\emph{et al}\onedot}
\def\eg{e.g\onedot}
\def\ie{i.e\onedot}
\def\etc{etc\onedot}
\makeatother

\begin{document}
\pagestyle{headings}
\mainmatter
\def\ECCVSubNumber{3129}  

\title{Improving Image Restoration by Revisiting Global Information Aggregation} 

\titlerunning{TLC: Test-time Local Converter}
%
\author{Xiaojie Chu \and
Liangyu Chen\textsuperscript{(\Letter)}\and
Chengpeng Chen \and
Xin Lu}
\authorrunning{Chu et al.}
%
\institute{MEGVII Technology\\
\email{\{chuxiaojie,chenliangyu\}@megvii.com,chencp@live.com,luxin941027@gmail.com}
}
\maketitle

\input{chaps/abstract}

\input{chaps/introduction}
\input{chaps/related_work}
\input{chaps/methods}
\input{chaps/experiments}
\input{chaps/conclusion}
\input{chaps/acknowledgements}
\clearpage
\input{chaps/supplementary}

\clearpage

\bibliographystyle{splncs04}

\input{3129.bbl}
\end{document}

%% file: chaps/abstract.tex
\begin{abstract}
Global operations, such as global average pooling, are widely used in top-performance image restorers. They aggregate global information from input features along entire spatial dimensions but behave differently during training and inference in image restoration tasks: they are based on different regions, namely the cropped patches (from images) and the full-resolution images. This paper revisits global information aggregation and finds that the image-based features during inference have a different distribution than the patch-based features during training. This train-test inconsistency negatively impacts the performance of models, which is severely overlooked by previous works. To reduce the inconsistency and improve test-time performance, we propose a simple method called Test-time Local Converter (TLC). Our TLC converts global operations to local ones only during inference so that they aggregate features within local spatial regions rather than the entire large images. The proposed method can be applied to various global modules (e.g., normalization, channel and spatial attention) with negligible costs. Without the need for any fine-tuning, TLC improves state-of-the-art results on several image restoration tasks, including single-image motion deblurring, video deblurring, defocus deblurring, and image denoising. In particular, with TLC, our Restormer-Local improves the state-of-the-art result in single image deblurring from 32.92 dB to 33.57 dB on GoPro dataset. The code is available at \url{https://github.com/megvii-research/tlc}.
\end{abstract}

%% file: chaps/introduction.tex
\section{Introduction}
Image restoration is the task of estimating the clean image from a corrupt (\eg, motion blur, noise, etc.) image. 
Recently, deep learning based models~\cite{fang2020multi,Zamir2021MPRNet,chen2021hinet} have achieved state-of-the-art (SOTA) performance in this field. 
The global information, which is aggregated along entire spatial dimensions, are increasingly indispensable for the top performance restorers: 
HINet~\cite{chen2021hinet} adopts Instance Normalization (IN~\cite{ulyanov2016instance}) module which performs global normalization along the entire spatial dimension. MPRNet~\cite{Zamir2021MPRNet}, SPDNet~\cite{fang2020multi}, FFA-Net~\cite{qin2020ffa}, etc. adopt Squeeze and Excitation (SE~\cite{hu2018squeeze}) module which learns to use global average-pooled features to selectively emphasise informative features. Restormer~\cite{zamir2021restormer} adopt \textit{transposed} self-attention for encoding the global information implicitly.

However, restoration models are usually trained on patches cropped from images and inference directly on full-resolution images~\cite{Zamir2021MPRNet,zamir2021restormer}. In contrast to resizing the input images during both training and inference in the high-level vision task, resizing the images in the low-level vision task is avoided to preserve the image details. As a result, the regional range of the inputs for training and inference varies widely. For example, during training in MPRNet~\cite{Zamir2021MPRNet}, the range of region for each patch is only 7\% of full-resolution images ($256 \times 256$ vs. $720 \times 1280$) in GoPro dataset. 
In this case, the model can only learn to encode a local part of the image due to the limited region of patches (Figure~\ref{fig:pipline_a}).
It may be difficult to encode the global clues of full-resolution images, thereby providing sub-optimal performance at test time. 
This potential issue is severely overlooked by previous works.

This paper revisits the global information aggregation in image restoration tasks. We analyze the global avg-pooled features and find that the entire-image-based features during inference may distribute very differently from the patch-based features during training (Figure~\ref{fig:distribution} Left). This shifts in the global information distribution in training and inference can negatively impact the performance of model.
To solve this issue, we proposed a novel test-time approach called Test-time Local Converter (TLC) for bridging the gap of information aggregation between training and inference. 
The global operation (\eg, global average pooling in SE module~\cite{hu2018squeeze}) is converted to a local one only during inference, so that they aggregate features within local spatial regions as in the training phase (Figure~\ref{fig:pipline_b}). As a result, the entire-image-based ``local'' information during inference has similar distribution as patches-based ``global'' information during training (Figure~\ref{fig:distribution} Right).
The proposed technique is generic in the sense that it can be applied on top of any global operation without any fine-tuning, and boost the performance of various modules (\eg, SE, IN) with negligible costs.

\input{tabs/Fig_pipeline}
Our TLC can be conveniently applied to already trained models.
We conduct extensive experiments to demonstrate the effectiveness of TLC over a variety of models and image restoration tasks. 
For example, for single-image motion deblurring on GoPro dataset~\cite{nah2017deep},
our TLC improves the PSNR of HINet~\cite{chen2021hinet}, MPRNet~\cite{Zamir2021MPRNet}, and Restormer~\cite{zamir2021restormer} by 0.37 dB, 0.65 dB, and 0.65 dB, respectively. 
Remarkably, TLC improves the state-of-the-art results on single-image motion deblurring, video motion deblurring, defocus deblurring (single-image and dual-pixel data), and image denoising (gaussian grayscale/color denoising).

Our contributions can be summarized as follows:
\begin{enumerate}
   \item To the best of our knowledge, we are the first to point out the inconsistency of the global information distribution in training (with cropped patches from images) and inference (with the full-resolution image) in image restoration tasks, which may harm model performance.

   \item To reduce the distribution shifts between training and inference, we propose Test-time Local Converter (TLC) that converts the region of feature aggregation from global to local only at test time. 
   Without retraining or fine-tuning, TLC significantly improves the performance of various modules with negligible costs by reducing the train-test inconsistency.
   
   \item Extensive experiments show that our TLC improves state-of-the-art results on various image restoration tasks.
\end{enumerate}

%% file: tabs/Fig_pipeline.tex
\begin{figure}[t]
     \centering
     \begin{subfigure}[b]{0.36\textwidth}
         \centering
         \includegraphics[width=\textwidth]{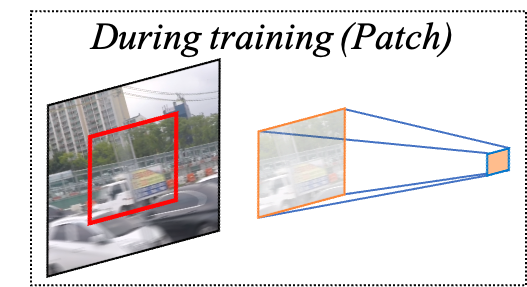}
         \caption{Global operation}
         \label{fig:pipline_a}
     \end{subfigure}
     \hfill
     \begin{subfigure}[b]{0.61\textwidth}
         \centering
         \includegraphics[width=\textwidth]{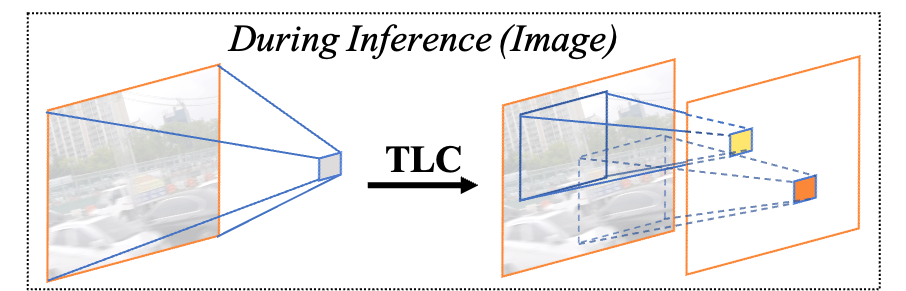}
         \caption{Test-time Local Converter (Ours)}
         \label{fig:pipline_b}
     \end{subfigure}
        \caption{{Illustration of global operation and our TLC}: (a) During training, limited by the cropped patches, global operation learns representation to local region in the original image; (b) During inference, global operation extract global representations based on full-resolution image. Our TLC convert the global operation to a local one so that it extract representations based on local spatial region of features as in training phase.}
        \label{fig:pipline}
\end{figure}

%% file: chaps/related_work.tex
\section{Related Work}
\subsubsection{Image Restoration.}
Image Restoration tasks, \eg denoising, deblurring, deraining, dehazing, \etc aim to restore the degraded image to the clean one. Deep learning based restoration models have achieved state-of-the-art results~\cite{fang2020multi,Zamir2021MPRNet,chen2021hinet,zamir2021restormer} recently. The training data are cropped into patches and fed into the model in the training phase. Most methods~\cite{qin2020ffa,fang2020multi,Zamir2021MPRNet} inference by the full-resolution image, which leads to a train-test inconsistency problem. Some methods~\cite{chen2021hinet,chen2021pre} divide the input image into patches with fixed size and process each patch independently, but this strategy may introduce boundary artifacts~\cite{lee2015block,liang2021swinir}.

\subsubsection{Global Information in Image Restoration Models.}
Attention modules are designed to model long-range dependency using a single layer directly. SENet~\cite{hu2018squeeze} and GENet~\cite{hu2018gather} reweight channel dependency with global information aggregated by global average pooling. CBAM~\cite{woo2018cbam} uses both avg-pooled and max-pooled features to rebalance the importance of different spatial positions and channels.
These channel and spatial attention modules have been successfully adopted to image restoration models for various tasks, \eg, deblurring~\cite{suin2020spatially,Zamir2021MPRNet,chen2022simple} deraining~\cite{li2018recurrent,fang2020multi}, super-resolution~\cite{zhang2018image,chu2022nafssr},  denoising~\cite{anwar2019real,Zamir2020CycleISP,zamir2021restormer} and dehazing~\cite{qin2020ffa}. 

Besides, HINet~\cite{chen2021hinet} introduces Instance Normalization (IN~\cite{ulyanov2016instance}) to image restoration tasks, which normalizes each channel of the features by its mean and variance. Once again, the performance improvement brought by IN proves the effectiveness of global information. 

This paper mainly discusses these modules, which aggregate information from all spatial positions in input features (\ie, globally), as representatives.
We find that the performance of these modules may be sub-optimal due to the train-test inconsistency mentioned above.

\subsubsection{Local Spatial Information Modules.}
In local spatial schemes, the information is computed within a local spatial area for each pixel. 
Local Response Normalization (LRN)~\cite{lyu2008nonlinear,jarrett2009best,krizhevsky2012imagenet} computes the statistics in a small neighborhood for each pixel. To reduce the computational loads, SwinIR~\cite{liu2021swin} and Uformer~\cite{wang2021uformer} apply self-attention within small spatial windows of size $8 \times 8$ around each pixel.
In semantic image synthesis tasks, SPatially-Adaptive (DE)normalization (SPADE)~\cite{park2019semantic} utilize the input semantic layout for modulating the activations through a spatially-adaptive, learned transformation.
Spatial region-wise normalization (RN)~\cite{yu2020region}, is proposed for better inpainting network training. 
  
However, directly applying those modules to existing restoration models is not practical, as retraining or finetuning are required. Besides, these modules are designed to model local context in both the training and inference phase that they are constrained to have limited sizes of receptive field.
Conversely, our proposed approach does not need to retrain or finetune the model. The region's size used for information aggregation during inference will be equal to or larger than the size of the input during training.

%% file: chaps/methods.tex
\newcommand{\FF}{\mathbf{F}}
\newcommand{\XX}{\mathbf{X}}
\newcommand{\KK}{\mathbf{K}}
\newcommand{\RR}{\mathbb{R}}
\newcommand{\UU}{\mathbf{U}}
\newcommand{\VV}{\mathbf{V}}
\newcommand{\WW}{\mathbf{W}}
\newcommand{\uu}{\mathbf{u}}
\newcommand{\vv}{\mathbf{v}}
\newcommand{\zz}{\mathbf{z}}
\newcommand{\xx}{\mathbf{x}}
\newcommand{\ThTh}{\boldsymbol{\theta}}

\section{Analysis and Approach}
In this section, we first introduce the image restoration pipeline and analyze the train-test inconsistency of global information aggregation induced by it.
Next, to solve the inconsistency, we illustrate our novel approach, Test-time Local Converter (TLC), and the details of extending TLC to existing modules.

\subsection{Revisit Global Operations in Image Restoration Tasks}
\subsubsection{Image Restoration Pipeline.}
We briefly describe the image restoration pipe-line used in the state-of-the-art methods. 
For practical application, datasets for image restoration tasks (\eg, deblurring) are usually composed of high-resolution images. 
Due to the need for data augmentation and the limitation of GPU memory, it is common practice to train models with small patches cropped from high-resolution images (Fig.~\ref{fig:pipline_a}). For example, MPRNet and HINet are trained on $256 \times 256$ patches cropped from $720 \times 1280$ images in GoPro datasets. During inference, the trained model directly restores high-resolution images (Fig.~\ref{fig:pipline_b} Left).
Therefore, there are train-test inconsistencies of the inputs to the model: a local region of the image during training and the entire image during inference.

%


\subsubsection{Train-Test Inconsistency of Global Information Aggregation. }
Unlike local operations (\eg, convolution) that operate within a local spatial area for each pixel, global operations (\eg, global average pool and global attention) operate along entire spatial dimensions. As a result, global operations have global receptive fields on arbitrary input resolutions. 

However, the range of receptive fields for global operation is limited by the size of input features.
This property introduces significantly different behaviors for global operations during training and inference in image restoration tasks: their input features are based on different range of regions, namely the cropped patches (from images) and the full-resolution images. 
This inconsistency will affect the generalization of models. In training, parameters are optimized by the patches-based features. While in the test phase, the layer inference the results by the entire-image-based features. 
In the following, we analyze the behavior of global information aggregation, both qualitatively and statistically. 

\subsubsection{Statistical Analysis.}
To analyze the effects of train-test inconsistency of global information, we compare the mean statistics based on patches in the training set and full-resolution images in the test set. The mean statistics (\ie, global average-pooled features) are aggregated by the first SE layer of the second encoder in MPRNet~\cite{Zamir2021MPRNet}. As shown in Fig.~\ref{fig:distribution}, the mean statistics distribution shifts from training (green) to inference (blue).
It is hard for restorers to adapt to the severe changes in information distribution, resulting in performance degradation.

\subsubsection{Qualitative Analysis.}
Intuitively, consistent with the training phase, cropping the images into patches and predicting the results independently during inference can alleviate the patch/full-image inconsistency issue described above. We conduct a visual comparison of the MPRNet deblurring results on GoPro datasets. 
Fig.~\ref{fig:deblurring} shows a challenging visual example.
The image-based result (Fig.~\ref{fig:deblurring}b) fails to remove blurs completely. On the contrary, the patch-based result (Fig.~\ref{fig:deblurring}c) is cleaner with less motion blur but introduces the artifacts at the patch boundaries. This confirms that direct inference on full-resolution results in sub-optimal performance. 
Though cropping images for inference improves the quality of image recovery, such a strategy will inevitably cause a new problem, \ie, patch boundary artifacts. 

\input{tabs/Fig_visualization}

\input{tabs/Fig_localsize}

\subsection{Test-time Local Converter}\label{sec:method:TLC}

In order to reducing train-test inconsistency and improve test-time performance of model, we propose a test-time solution named Test-time Local Converter (TLC).
Instead of changing the training strategy or cropping the images, our TLC directly change the range of region for information aggregation at feature level during inference phase. 
As shown in Figure~\ref{fig:pipline_b}, TLC converts the spatial information aggregation operation from global to local, \ie, each pixel of the feature aggregates its feature locally.
In detail, the input feature $\mathbf{X}$ of global operation is sliced into overlapping window with size of $K_h \times K_w$ (which are treated as hyper-parameters). Then, information aggregation operation is applied independently to each overlapping window. 
As a result, the statistics distribution shifts are reduced by TLC as shown in Fig.~\ref{fig:distribution}: the statistics distribution obtained by our MPRNet-Local (red) is close to the original MPRNet in the training phase (green). Besides, as shown in Fig.~\ref{fig:deblurring}d, our TLC generates a sharp image without boundary artifacts.

An advantage of our intentionally simple design is that efficient implementations of local processing make extra computation cost negligible, allowing image restorers to use TLC feasibly. Next, we will discuss the implementations of average operation, which is an example of information aggregation and is widely used in models for image restoration. 

\subsubsection{Efficient Implementation of Information Aggregation.} 
The (global) information aggregation of a feature layer $\mathbf{X} \in \mathbb{R}^{H \times W}$ (without loss of generality, we ignore the channel dimension), can be formulated as:
\begin{equation}
\label{eqn:global_agg}
\Phi(\mathbf{X}, f)=\frac{1}{HW}\sum_{p=1}^{H} \sum_{q=1}^{W} f(\mathbf{X}_{p, q}).
\end{equation}
where  $f: \mathbb{R} \to \mathbb{R}$ defines how information are calculated, and $\Phi(\XX, f) \in \mathbb{R}$ denotes the aggregated information. It's computational complexity is $\mathcal{O}(HW)$.
For local information aggregation, each pixel $\eg (i,j)$ aggregates the information in a local window (size $K_h \times K_w$) of feature $\XX \in \mathbb{R}^{H\times W}$ could be formulated as:
\begin{equation}
\mathbf{\Psi}(\mathbf{X}, f)_{i,j}=\frac{1}{K_hK_w}\sum_p\sum_q f(\mathbf{X}_{p,q}), 
\end{equation}
where $(p,q)$ in the local window of $(i,j)$, $\mathbf{\Psi}(\mathbf{X}, f) \in \mathbb{R}^{H\times W}$ indicates the aggregated local information, and $K_h, K_w$ are hyperparameters. 

The edge case, eg. $(i,j)$ is the boundary of $\XX$, is not considered above for simplicity. In practice, we implement $\mathbf{\Psi}(\mathbf{X}, f)$ by two steps. First, sliding windows (size of $K_h \times K_w$) with stride equals $1$ to aggregate the local information for each pixel in non-edge case. Second, padding the result by replication of its boundary for edge case. The first step's computational complexity is $\mathcal{O}(HWK_hK_w)$. 
But mean/sum aggregation within each local window could be treated as \textit{submatrix sum} problem and solved by prefix sum trick~\cite{harris2007parallel} with $\mathcal{O}(1)$ complexity~\cite{amir2004submatrices}.
As a result, the overall complexity could be reduced to $\mathcal{O}(HW)$ which is consistent with global information aggregation operation, \ie Eq.(\ref{eqn:global_agg}). 
Therefore, our TLC do not induce a computational bottleneck. 

\subsection{Extending TLC to Existing Modules}\label{sec:method:extend}
In this subsection, we borrow the notations defined above (\eg,  $\Phi$/$\mathbf{\Psi}$ denotes global/local information aggregation operation, respectively). To extend TLC to existing modules, we convert the information aggregation operation from global (\ie, $\Phi$) to local (\ie, $\mathbf{\Psi}$). 
In the following, we take Squeeze-and-Excitation(SE) and Instance Normalization(IN) as representatives, and it can be easily applied to other normalization modules such as Group Normalization (GN~\cite{wu2018group}) or variants of SE (\eg CBAM~\cite{woo2018cbam}, GE~\cite{hu2018gather}).

\textbf{Extending TLC to SE Block.} 
We briefly revisit the squeeze-and-excitation (SE~\cite{hu2018squeeze}) block first. For a feature map $\XX \in \mathbb{R}^{H\times W \times C}$ with a spatial size of $(H,W)$ and $C$ channels, SE block first squeezes the global spatial information into channels, it could be denoted as $\Phi(\XX^{(c)}, id), \forall c \in [C]$, where $id(t) = t, \forall t \in \mathbb{R}$. And then, a multilayer perceptron~(MLP) follows to evaluate the channel attention, which re-weights the feature map. 
The squeeze on the global spatial dimension could be sub-optimal as global information distribution shifts. To solve this, we extend TLC to SE by replacing $\Phi(\XX^{(c)}, id)$ to $\mathbf{\Psi}(\XX^{(c)}, id), \forall c \in [C]$. As in SE, an MLP along the channel dimension follows. Differently, the feature map is re-weighted by the element-wise attention in this case. 

\textbf{Extending TLC to IN.}
For a feature map $\XX \in \mathbb{R}^{H\times W}$ (we omit the channel dimension for simplicity), the normalized feature $\mathbf{Y}$ by IN is computed as:  $\mathbf{Y}={(\mathbf{X}-\mu)}/{\sigma}$,
where statistics $\mu$ and $\sigma$ are the mean and variance computed over the global spatial of $\mathbf{X}$:
\begin{equation}
\label{eqn:mu_sigma_by_phi}
\begin{split}
\mu=\Phi(\mathbf{X}, id), ~\sigma^2=\Phi(\mathbf{X}, sq)-\mu^2,
\end{split}
\end{equation}
where $id(t)=t, sq(t)=t^2, \forall t \in \mathbb{R}$. Besides, learnable parameters $\gamma, \beta$ are used to scale and shift the normalized feature $\mathbf{Y}$, we omit them for simplicity.
During inference, we can extend TLC to IN by replacing $\Phi(\XX, id)$ and $\Phi(\XX, sq)$ in Eq.(\ref{eqn:mu_sigma_by_phi}) to $\mathbf{\Psi}(\mathbf{X}, id)$ and $\mathbf{\Psi}(\mathbf{X}, sq)$ respectively. As a result, each pixel is normalized by statistics in neighborhood.

\textbf{Extending to transposed self-attention.}
As introduced in Sec. 3.2, our TLC can convert the \textit{transposed} self-attention in Restormer~\cite{zamir2021restormer} from global to local regions. However, due to inefficiency and limitation of GPU memory, \ie, different attention map for each pixel, we use a large stride rather than one to the TLC in transposed self-attention. Specifically, transposed self-attention is applied independently to each overlapping windows of $K_h\times K_w$ sliced from input features. The overlapping outputs are then fused by concatenating along spatial dimensions and averaging over the overlapping regions.  

\subsection{Discussion}
Apart from our method, the range of input at image level also has a direct impact on train-test inconsistency. On the one hand, larger size of patches used for training pushes the patch-based information closer to image-based information. On the other, dividing the image into patches for inference can avoid the patch/entire-image inconsistency. 
We will discuss these two possible solutions and their drawbacks next.

\textbf{Dividing the image into patches for independently inference}  may alleviate the inconsistency issue. However, such a strategy inevitably gives rise to two drawbacks~\cite{liang2021swinir}. First, border pixels cannot utilize neighbouring pixels that are out of the patch for image restoration. Second, the restored image may introduce ``boundary artifacts''~\cite{lee2015block} around each patch. 
As shown in Fig.~\ref{fig:deblurring}c, an obvious vertical split line is introduced by patch partition which severely damages the image quality. In contrast, our ``partition'' is at feature level instead of image level so that different local windows can still interact with each other through other modules (\eg, convolutions) in the network. As shown in Fig.~\ref{fig:deblurring}d, the proposed method generates much clearer images without artifacts.

Besides, inference with overlapping patches will introduce considerable additional computational costs, as the overlapping regions are restored twice or more by the entire model. While models with our TLC directly restore whole images and TLC has low extra computing costs (Table~\ref{tab:Unet_local}). 
Furthermore, boundary artifacts are also found in the predictions based on overlapping patches. 
More details of discussion and comparison are in the supplemental material.

\textbf{Training on full-images} instead of patches is another straightforward idea to bridge the gap in global information distribution between training and inference, but it is impracticable due to limited device constrains. The scaling up of resolution leads to prohibitively high GPU memory consumption with existing image restorers.
For example, using V100-32G, the size of patches for training  Restormer~\cite{zamir2021restormer} can only up to $384\times384$,
which is still significantly smaller than the original image size (\eg, $720\times1280$ in GoPro dataset). Furthermore, though Restormer is trained with larger patches than common practice, our TLC can significantly improve its performance (Table~\ref{tab:SOTA_deblur}).

%% file: tabs/Fig_visualization.tex
\begin{figure*}[t]
\centering

\includegraphics[width=\linewidth]{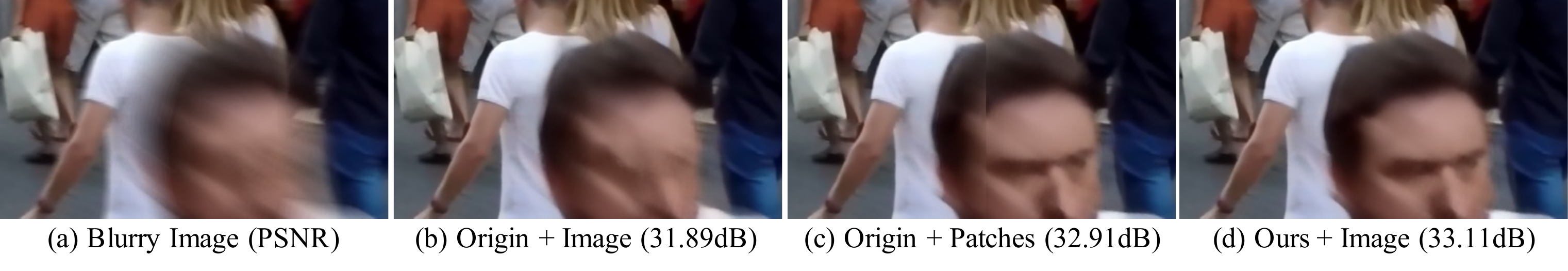}
\caption{
Visual comparison with different test-time methods to MPRNet~\cite{zamir2021multi} for image deblurring.
(a) Blurry image;
(b) Inference with image;
(c) Inference with cropped patches;
(d) Ours: TLC is adopted and inference based on images. 
Our TLC generates sharp image without boundary artifacts in (b).
}
\label{fig:deblurring}
\end{figure*}

%% file: tabs/Fig_localsize.tex
\begin{figure}[t]
\centering
\begin{subfigure}[b]{0.48\columnwidth}
\centering
\includegraphics[width=0.95\linewidth]{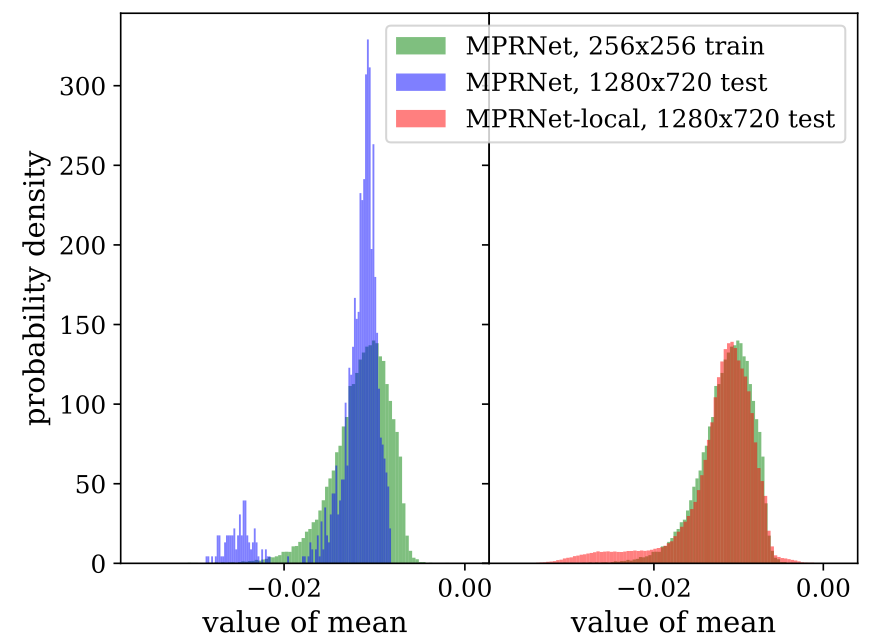}
\caption{
TLC can reduce distribution shifts (in red) caused by the inconsistency between training and testing (green vs. blue).
}
\label{fig:distribution}
\end{subfigure}
\hfill
\begin{subfigure}[b]{0.50\columnwidth}
\centering
\includegraphics[width=\linewidth]{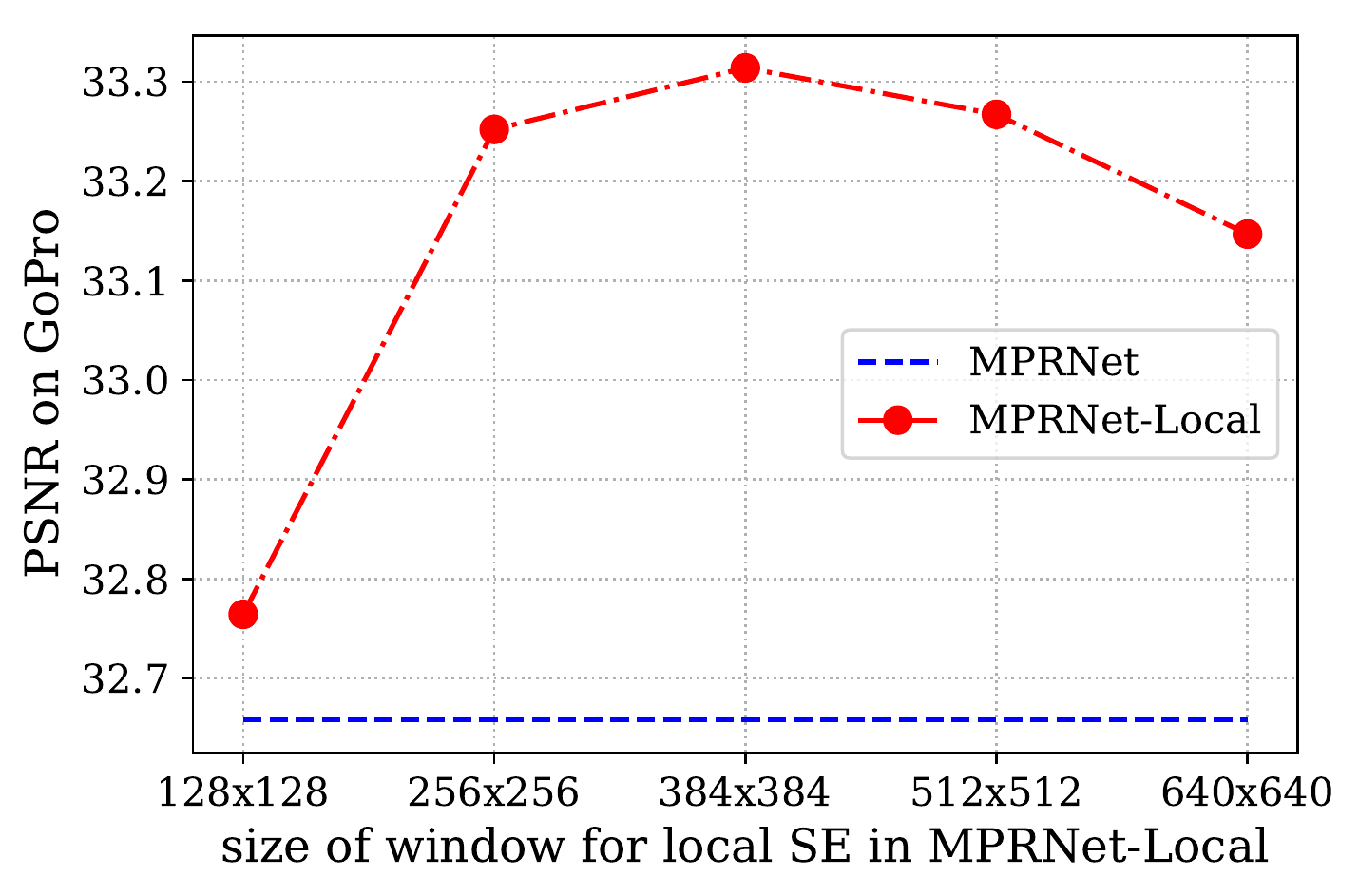}
\caption{TLC can significantly improve the performance of MPRNet over a wide range of hyperparameters (\ie, size of local window).
}
\label{fig:local_size}
\end{subfigure}
\vspace{-2mm}
\caption{
The effectiveness of TLC on MPRNet~\cite{Zamir2021MPRNet} (denoted as MPRNet-Local).
}
\label{fig:effectiveness_tlsc}
\end{figure}

%% file: chaps/experiments.tex
\section{Experiments}
In this section, we do quality and quantity experiments to show the effects of train-test inconsistency, and our proposed approach Test-time Local Converter (TLC) can reduce this inconsistency. Next, the extensibility of TLC and the choice of hyperparameters are discussed.

\subsection{Main Results}
To verify the effectiveness of the proposed TLC, we apply it to various existing top-performing models for six image restoration tasks: (1) single-image motion deblurring, (2) video deblurring, (3) defocus deblurring, (4) image denoising, (5) image deraining and (6) image dehazing.  We report the standard metrics in image restoration, including Peak Signal to Noise Ratio (PSNR) and Structural SIMilarity index (SSIM). 

\textbf{Implementation Details.}
We use the publicly available trained models with global operations (\eg, global attention, global normalization) and directly apply proposed TLC to them without any extra training. 
Specifically, for Restormer~\cite{zamir2021restormer} with TLC, the forward pass of transposed attention is applied independently to each overlapping window sliced from original input features. While for SE~\cite{hu2018squeeze} (used in MPRNet~\cite{Zamir2021MPRNet}, RNN-MBP~\cite{zhu2021deep}, SPDNet~\cite{fang2020multi} and FFANet~\cite{qin2020ffa}) and Instance Normalization~\cite{ulyanov2016instance} (used in HINet~\cite{chen2021hinet}), TLC is extended to them as illustrated in Sec.~\ref{sec:method:extend}.
Models with our TLC is marked with ``-Local'' suffix and the local window size is set to $384 \times 384$ if not specified. We will discuss the impact of this hyper-parameter in the Sec.~\ref{sec:local_size}.

\input{tabs/Tab_deblur}
\input{imgs/Fig_deblurring}
\textbf{Single-image Motion Deblurring.} We integrate our TLC with existing top-performing models (\eg HINet~\cite{chen2021hinet}, MPRNet~\cite{Zamir2021MPRNet}, and Restormer~\cite{zamir2021restormer}) and evaluate them on test set of GoPro~\cite{nah2017deep} and HIDE~\cite{shen2019human} dataset. 
As shown in Table~\ref{tab:SOTA_deblur}, the performance of both three models are significantly improved by our approach and our models achieve new state-of-the-art results on all datasets.

In detail, the PSNR on GoPro of HINet, MPRNet and Restormer are improved by 0.37 dB, 0.65 dB and 0.65dB, respectively. And our Restormer-local exceeds the previous best result~(\ie, Restormer~\cite{zamir2021restormer}) by 0.65 dB.
The PSNR on HIDE of HINet, MPRNet and Restormer are improved by 0.33 dB, 0.23 dB and 0.27 dB, respectively. And our Restormer-local exceeds the previous best result~(\ie, Restormer~\cite{zamir2021restormer}) by 0.27 dB. 
Visual results of our methods are shown in Fig.~\ref{fig:visual_deblurring}. As one can see, based on its significant quantitative improvements, TLC can help the original model generate more sharp images with clearer numeric symbols.

\input{tabs/Tab_deblur_v}
\textbf{Video Motion Deblurring.} We apply our TLC to state-of-the-art video deblurring method (\ie, RNN-MBP~\cite{zhu2021deep}) and evaluate different video deblurring algorithms on GoPro datasets. As shown in Table~\ref{tab:deblur_v}, our TLC improve previous state-of-the-art method by 0.48 dB on PSNR, and set a new state-of-the-art result at 33.80 dB.

\input{tabs/Tab_defocus}

\textbf{Defocus Deblurring.} Table~\ref{table:defocus} shows image fidelity scores of different defocus deblurring methods on the DPDD dataset~\cite{abuolaim2020defocus}.
Following~\cite{abuolaim2020defocus}, results are reported on traditional signal processing metrics (\ie, PSNR, SSIM, and MAE) and learned perceptual image patch similarity (LPIPS) proposed by~\cite{zhang2018unreasonable}.
TLC are applied to the state-of-the-art method Restormer and get Restormer-Local.
Our Restormer-Local significantly outperforms the state-of-the-art schemes for the single-image and dual-pixel defocus deblurring tasks on all scene categories.
Take PSNR as evaluation metrics, our TLC improves Restormer by 0.21$\sim$0.3 dB and 0.35$\sim$0.40 dB on single-image and dual-pixel defocus deblurring, respectively. 
Fig.~\ref{fig:dualpixel_deblurring} shows that our model recovered images of better quality on texture and edge detail.
\input{imgs/Fig_defocus}

\input{tabs/Tab_denoise}
\textbf{Image Denoise.} We perform denoising experiments on synthetic benchmark dataset Urban100~\cite{huang2015single} generated with additive white Gaussian noise.
Table~\ref{table:colordenoising} and Table~\ref{table:colordenoising} show PSNR scores of different approaches on several benchmark datasets for grayscale and color image denoising, respectively. Consistent with existing methods~\cite{zhang2021plug,liang2021swinir,zamir2021restormer}, we include noise levels 15, 25 and 50 in testing. The evaluated methods are divided into two experimental categories: (1) learning a single model to handle various noise levels, and (2) learning a separate model for each noise level. We apply TLC to state-of-the-art method Restormer. Our TLC brings 0.06$\sim$0.12 dB improvement on grayscale image denoising and brings 0.08$\sim$0.15 dB improvement on color image denoising. Fig.~\ref{fig:denoise} shows that our method clearly removes noise while maintaining fine details.
\input{imgs/Fig_denoise}

\textbf{Image Deraining.}
We compare the deraining results of SPDNet~\cite{fang2020multi} and our SPDNet-local on SPA-Data~\cite{wang2019spatial} benchmark.
As shown in Table~\ref{tab:SOTA.derain}, our approach brings 0.18 dB improvement to SPDNet. Fig.~\ref{fig:deraining} shows that our approach recovered images of better quality on both details and color fidelity.

\textbf{Image Dehazing.}
We compare the dehazing results of FFANet~\cite{qin2020ffa} and our FFANet-local on Synthetic Objective Testing Set (SOTS) from RESIDE~\cite{li2018benchmarking} dataset.
The local window size is set to $416 \times 416$. 
As shown in Table~\ref{tab:SOTA.dehaze}, our approach brings 0.42 dB improvement to FFANet in outdoor scenarios.  We also test the results on realistic hazy images in RESIDE~\cite{li2018benchmarking} dataset for subjective assessment. As shown in Fig.~\ref{fig:dehazing}, our FFANet-Local effectively removes hazy and generate visually pleasing result with high color fidelity. More high-resolution visualization results are in the supplemental material.

\input{tabs/Tab_restoration}
\input{imgs/Fig_derain}

\input{imgs/Fig_dehazing}

\subsection{Size of Local Window}~\label{sec:local_size} 
Size of local window is a hyper-parameter for TLC, which controls the scope of local information aggregation operation. To determinate the hyperparameter, \ie, local window size of each layer (which aggregates the spatial context) as we mentioned in ~Sec.\ref{sec:method:TLC}, we propose a simple strategy:
A calibration image is fed into the model, and the spatial sizes of these feature layers are recorded as their local window size. Therefore, the hyperparameter could be determined by the spatial size of the calibration image, and we denoted the image size as ``local window size'' for simplicity in the following. Besides, the calibration could be accomplished offline, thus does not increase the test latency. 

We apply TLC on MPRNet~\cite{Zamir2021MPRNet} to investigate the impact of different size of local window on the model performance. As shown in Fig.~\ref{fig:local_size}, TLC can significantly improve the performance of MPRNet over a wide range of window size (from $256\times256$ to $640\times640$).
Interestingly, the optimal window size (\ie, $384 \times 384$) for the test phase is not exactly equal but may be larger than the training patch size (\ie, $256\times256$). We conjecture this is caused by the trade-off between the benefits of more information provided by the larger window and the side-effects of statistic inconsistency between training and inference.
In addition, since our approach does not require retraining, it is easy and flexible to tune the size of local window.

\subsection{Extensibility and Complexity} \label{sec:ext_unet}
\input{tabs/Fig_local}
We apply TLC to various modules and compare its improvement of performance and complexity of computation.
We use a simple UNet model (\ie, HINet Simple without HIN~\cite{chen2021hinet}) as baseline (denoted as UNet).
Attention modules (\eg, SE~\cite{hu2018squeeze}, GE-$\theta^-$~\cite{hu2018gather} and CBAM~\cite{woo2018cbam}) are added to UNet encoder following SENet~\cite{hu2018squeeze}, while Normalization modules (\eg, IN~\cite{ulyanov2016instance} and GN~\cite{wu2018group}) are added to the UNet following HINet~\cite{chen2021hinet}. 

\textbf{Implementation details.}
Models are trained on GoPro~\cite{nah2017deep} dataset following the most training detail of HINet Simple~\cite{chen2021hinet}. 
Specially, the default size of patches for training is $256\times256$, and the default batch size is 64. We also use warm-up strategy in the first 5000 iterations. According to Sec.~\ref{sec:local_size}, the local window size is set to $384 \times 384$ during inference.
We use MACs (\ie multiplier-accumulator operations) to evaluate the computational cost of models, which is estimated when the input is $ 512\times 512$. 

\textbf{Results.} 
As shown in Table~\ref{tab:Unet_local}, our approach achieves performance gains with marginal costs. In detail, TLC improves the performance (\ie, PSNR) of IN, GN and GE-$\theta^-$ by 0.16 dB, 0.12 dB, and 0.17 dB, respectively. For SE and CBAM, TLC boosts the performance(\ie PSNR) by 0.39 dB and 0.52 dB respectively with less than 0.2\% extra MACs. It demonstrates the extensibility, effectiveness and efficiency of TLC.

%% file: tabs/Tab_deblur.tex
\begin{table}[t]
\caption{{Image motion deblurring} comparisons on GoPro~\cite{nah2017deep} and HIDE~\cite{shen2019human}}
\label{tab:SOTA_deblur}
\centering
\resizebox{\linewidth}{!}{
\begin{tabular}{c|c|cccccccc}
\toprule
 & {Method} & Gao~\etal              & DBGAN                       & MT-RNN                & DMPHN                 & Suin~\etal                      & SPAIR                        & MIMO-UNet+                & IPT                 \\ 
{Dataset} &                         & ~\cite{gao2019dynamic} & ~\cite{zhang2020deblurring} & ~\cite{park2020multi} & ~\cite{zhang2019deep} & ~\cite{suin2020spatially} & ~\cite{purohit2021spatially} & ~\cite{cho2021rethinking} & ~\cite{chen2021pre} \\ \midrule
\multirow{2}{*}{{GoPro}}   & PSNR$\uparrow$                    & 30.90                  & 31.10                       & 31.15                 & 31.20                 & 31.85                     & 32.06                        & 32.45                     & 32.52               \\
                         & SSIM$\uparrow$                    & 0.935                  & 0.942                       & 0.945                 & 0.940                 & 0.948                     & 0.953                        & 0.957                     & -                   \\ \midrule
\multirow{2}{*}{{HIDE}}    & PSNR$\uparrow$                    & 29.11                  & 28.94                       & 29.15                 & 29.09                 & 29.98                     & 30.29                        & 29.99                     & -                   \\
                         & SSIM$\uparrow$                    & 0.913                  & 0.915                       & 0.918                 & 0.924                 & 0.930                     & 0.931                        & 0.930                     & -                                   \\ \bottomrule 
\end{tabular}}
\resizebox{\linewidth}{!}{
\begin{tabular}{c|c|cc|cc|cc}
\toprule
& {Method} & ~HINet~                 & {HINet-Local}~     & ~MPRNet~                  & {MPRNet-Local}~              & ~Restormer~                  & {Restormer-Local}~ \\
{Dataset}  &                         & ~\cite{chen2021hinet} & {(Ours)}          & ~\cite{Zamir2021MPRNet} & {(Ours)}                    & ~\cite{zamir2021restormer} & {(Ours)}          \\ \midrule
\multirow{2}{*}{{GoPro}}   & PSNR$\uparrow$                    & 32.71                 & {33.08$^{+0.37}$} & 32.66                   & {33.31$^{+0.65}$}    & 32.92                      & \textbf{33.57$^{+0.65}$}   \\
                         & SSIM$\uparrow$                    & 0.959                 & {0.962$^{+0.003}$}  & 0.959                   & {0.964$^{+0.005}$} & 0.961                      & \textbf{0.966$^{+0.005}$}  \\ \midrule
\multirow{2}{*}{{HIDE}}    & PSNR$\uparrow$                    & 30.33                 & {30.66$^{+0.33}$}   & 30.96                   & {31.19$^{+0.23}$}    & 31.22                      & \textbf{31.49$^{+0.27}$}   \\
                         & SSIM$\uparrow$                    & 0.932                 & {0.936$^{+0.004}$}  & 0.939                   & {0.942$^{+0.003}$}   & 0.942                      & \textbf{0.945$^{+0.003}$}  \\\bottomrule
\end{tabular}
}
\end{table}

%% file: imgs/Fig_deblurring.tex
\begin{figure}[!t]
\centering
\scalebox{0.98}{
\begin{tabular}[b]{c@{ } c@{ }  c@{ } c@{ }}
    \multicolumn{2}{c}{\multirow{4}{*}{\includegraphics[width=.475\textwidth,valign=t]{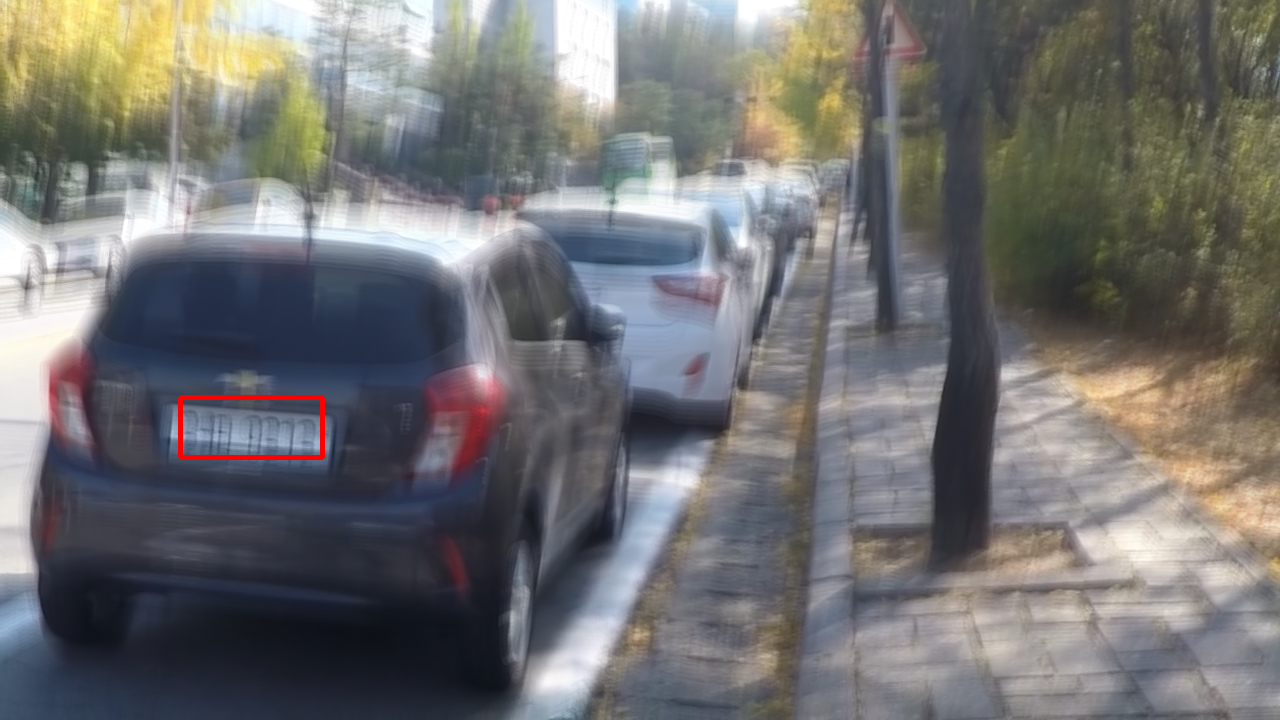}}} &   
    \includegraphics[clip,width=.24\textwidth,valign=t]{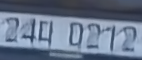}&
  	\includegraphics[clip,width=.24\textwidth,valign=t]{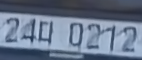}
\\
  & &  \small~23.71 dB &\small~25.88 dB   \\
    
  & & \small~HINet~\cite{chen2021hinet} & \small~HINet-Local    \\

  & &
     \includegraphics[clip,width=.24\textwidth,valign=t]{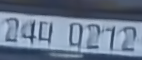}&  
     \includegraphics[clip,width=.24\textwidth,valign=t]{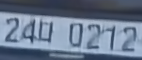}\\

    \multicolumn{2}{c}{ \small~21.38 dB }& \small~24.02 dB 
     & \small~28.53 dB \\
      \multicolumn{2}{c}{\small~Blurry Image}   & \small MPRNet~\cite{Zamir2021MPRNet} & \small MPRNet-Local \\
\includegraphics[clip,width=.24\textwidth,valign=t]{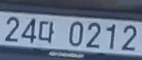} & \includegraphics[clip,width=.24\textwidth,valign=t]{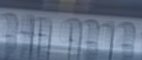} &  
      \includegraphics[clip,width=.24\textwidth,valign=t]{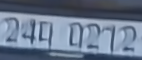}&
     \includegraphics[clip,width=.24\textwidth,valign=t]{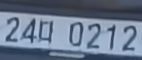}
\\
\small~PSNR & \small~21.38 dB  & \small 25.36 dB  & \small 29.00 dB \\
  \small~Reference & \small~Blurry& \small~Restormer~\cite{zamir2021restormer} & \small~Restormer-Local \\
\end{tabular}}
\caption{Qualitative evaluation of our TLC on single image motion deblurring methods. 
Models with our TLC (denoted with -Local suffix) generates sharper result than original ones. 
} \label{fig:visual_deblurring}
\end{figure}


%% file: tabs/Tab_deblur_v.tex
\begin{table}[t]
\caption{{Video deblurring} comparisons on the GoPro~\cite{nah2017deep} dataset}
\label{tab:deblur_v}
\resizebox{\linewidth}{!}{
\begin{tabular}{c|ccccccc|cc}
\toprule
{Method} & SFE   & IFI-RNN & ESTRNN & EDVR  & TSP    & PVDNet & GSTA   & RNN-MBP & {RNN-MBP-Local} \\
   & \cite{xiang2020deep}                 &  \cite{nah2019recurrent}     &   \cite{zhong2020efficient}      &  \cite{wang2019edvr}      &  \cite{pan2020cascaded}     &   \cite{son2021recurrent}     &     \cite{suin2021gated}   &    \cite{zhu2021deep}     &   {(Ours)}            \\ \midrule
PSNR$\uparrow$                    & 31.01 & 31.05   & 31.07  & 31.54 & 31.67  & 31.98  & 32.10  & 33.32   & \textbf{33.80$^{+0.48}$}         \\
SSIM$\uparrow$                    & 0.913 & 0.911   & 0.902 & 0.926 & 0.928 & 0.928 & 0.960 & 0.963  & \textbf{0.966$^{+0.003}$}       \\
\bottomrule
\end{tabular}}
\end{table}

%% file: tabs/Tab_defocus.tex
\begin{table*}[!t]
\centering
\caption{{Defocus deblurring} comparisons on the DPDD testset~\cite{abuolaim2020defocus} (containing 37 indoor and 39 outdoor scenes). S: single-image defocus deblurring. D: dual-pixel defocus deblurring. Our Restormer-Local sets new state-of-the-art for both single-image and dual pixel defocus deblurring 
}
\label{table:defocus}
\resizebox{\linewidth}{!}{
\begin{tabular}{l | c c c c | c c c c | c c c c }
\toprule
   & \multicolumn{4}{c|}{\textbf{Indoor Scenes}} & \multicolumn{4}{c|}{\textbf{Outdoor Scenes}} & \multicolumn{4}{c}{\textbf{Combined}} \\
\cline{2-13}
   \textbf{Method} & PSNR$\textcolor{black}{\uparrow}$ & SSIM$\textcolor{black}{\uparrow}$& MAE$\textcolor{black}{\downarrow}$ & LPIPS$\textcolor{black}{\downarrow}$  & PSNR$\textcolor{black}{\uparrow}$ & SSIM$\textcolor{black}{\uparrow}$& MAE$\textcolor{black}{\downarrow}$ & LPIPS$\textcolor{black}{\downarrow}$  & PSNR$\textcolor{black}{\uparrow}$ & SSIM$\textcolor{black}{\uparrow}$& MAE$\textcolor{black}{\downarrow}$ & LPIPS$\textcolor{black}{\downarrow}$   \\
\hline
EBDB$_S$~\cite{chen2021pre} & 25.77 & 0.772 & 0.040 & 0.297 & 21.25 & 0.599 & 0.058 & 0.373 & 23.45 & 0.683 & 0.049 & 0.336 \\
DMENet$_S$~\cite{lee2019deep}  & 25.50 & 0.788 & 0.038 & 0.298 & 21.43 & 0.644 & 0.063 & 0.397 & 23.41 & 0.714 & 0.051 & 0.349 \\
JNB$_S$~\cite{shi2015just} & 26.73 & 0.828 & 0.031 & 0.273 & 21.10 & 0.608 & 0.064 & 0.355 & 23.84 & 0.715 & 0.048 & 0.315 \\
DPDNet$_S$~\cite{abuolaim2020defocus} &26.54 & 0.816 & 0.031 & 0.239 & 22.25 & 0.682 & 0.056 & 0.313 & 24.34 & 0.747 & 0.044 & 0.277\\
KPAC$_S$~\cite{son2021single} & 27.97 & 0.852 & 0.026 & 0.182 & 22.62 & 0.701 & 0.053 & 0.269 & 25.22 & 0.774 & 0.040 & 0.227 \\
IFAN$_S$~\cite{lee2021iterative} & 28.11 & 0.861 & 0.026& 0.179 & 22.76 & 0.720& 0.052 & 0.254 & 25.37& 0.789& 0.039& 0.217\\ \hline
Restormer$_S$~\cite{zamir2021restormer}&   28.87 &   0.882 &   0.025&   0.145&   23.24 &   0.743 &   0.050&   0.209 &   25.98 &   0.811 &   0.038 &   0.178  \\
\textbf{Restormer-Local$_S$}&   \textbf{29.08}&\textbf{0.888}&\textbf{0.024}&\textbf{0.139}&\textbf{23.54}&\textbf{0.765}&\textbf{0.049}&\textbf{0.195}&\textbf{26.24}&\textbf{0.825}&\textbf{0.037}&\textbf{0.168}   \\
\midrule
\midrule
DPDNet$_D$~\cite{abuolaim2020defocus} & 27.48 & 0.849 & 0.029 & 0.189 & 22.90 & 0.726 & 0.052 & 0.255 & 25.13 & 0.786 & 0.041 & 0.223 \\
RDPD$_D$~\cite{abuolaim2021learning} & 28.10 & 0.843 & 0.027 & 0.210 & 22.82 & 0.704 & 0.053 & 0.298 & 25.39 & 0.772 & 0.040 & 0.255 \\
Uformer$_D$~\cite{wang2021uformer} & 28.23 & 0.860 & 0.026 & 0.199 & 23.10 & 0.728 & 0.051 & 0.285 & 25.65 & 0.795 & 0.039 & 0.243 \\
IFAN$_D$~\cite{lee2021iterative} & 28.66& 0.868& 0.025& 0.172& 23.46& 0.743& 0.049& 0.240& 25.99& 0.804& 0.037& 0.207\\ \hline
Restormer$_D$~\cite{zamir2021restormer}&   29.48  &   0.895 &   0.023&   0.134&   23.97 &   0.773 &   0.047&   0.175 &   26.66 &   0.833 &   0.035 &   0.155\\
\textbf{Restormer-Local$_D$}&   \textbf{29.83}&\textbf{0.903}&\textbf{0.022}&\textbf{0.120}&\textbf{24.37}&\textbf{0.794}&\textbf{0.045}&\textbf{0.159}&\textbf{27.02}&\textbf{0.847}&\textbf{0.034}&\textbf{0.140}  \\
\bottomrule
\end{tabular}}
\end{table*}

%% file: imgs/Fig_defocus.tex
\begin{figure*}[!t]
\centering
\scalebox{0.99}{
\begin{tabular}[b]{c@{ } c@{ }  c@{ } c@{ } c@{ }   }
\includegraphics[width=.259\textwidth,valign=t]{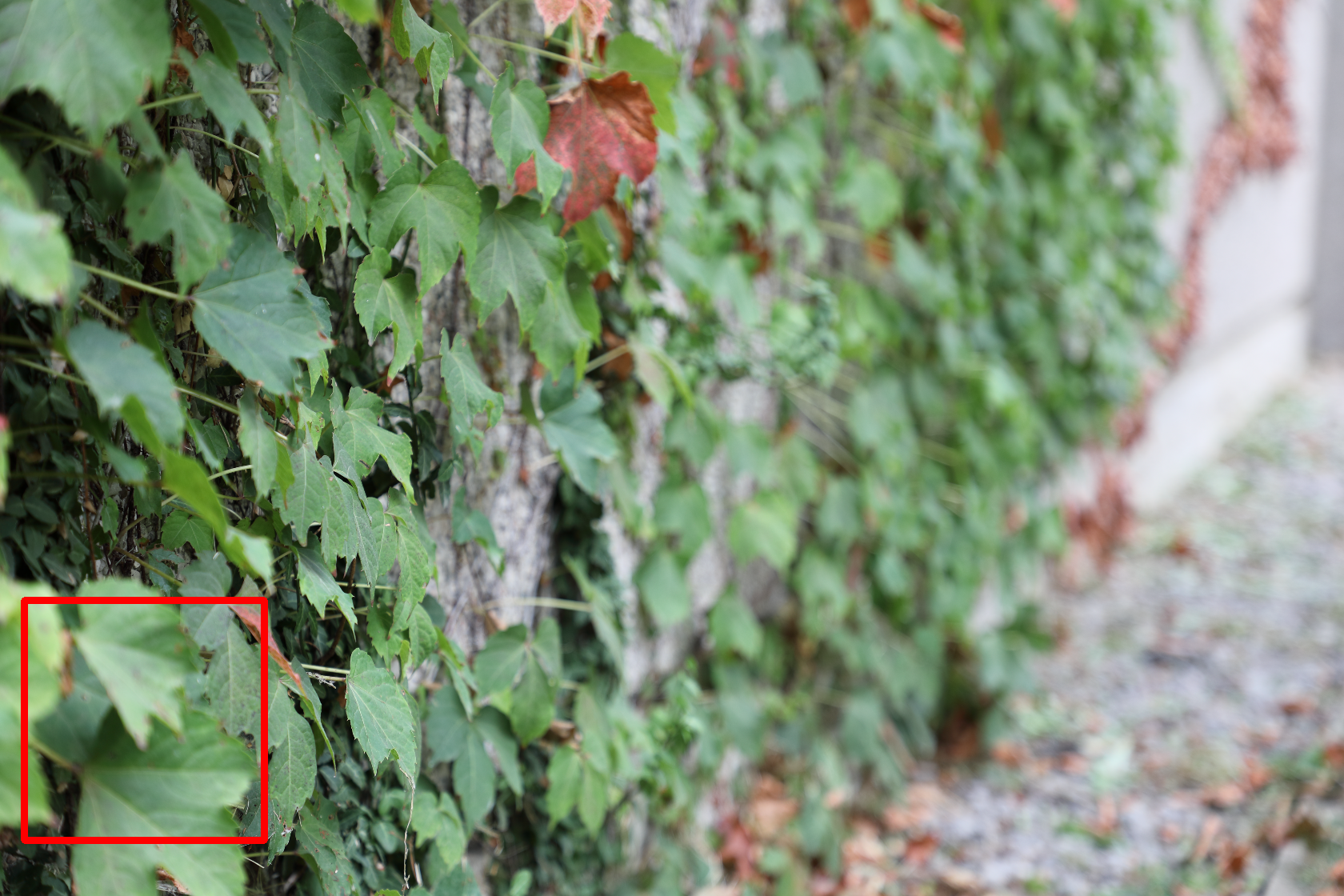}& 
    \includegraphics[width=.172\textwidth,valign=t]{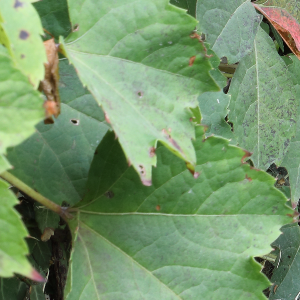} &
    \includegraphics[width=.172\textwidth,valign=t]{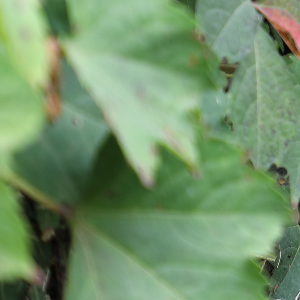} &
    \includegraphics[width=.172\textwidth,valign=t]{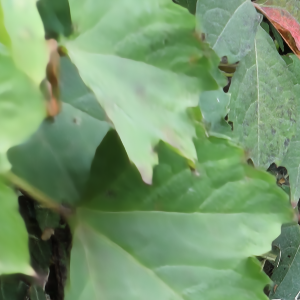} &
    \includegraphics[width=.172\textwidth,valign=t]{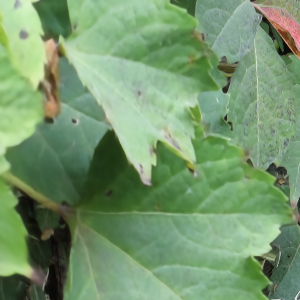}
\\
\scriptsize~20.72 dB  &  \scriptsize~PSNR &\scriptsize~20.72 dB  & \scriptsize~25.88 dB & \scriptsize~27.16 dB   \\
   \scriptsize~Blurry Image & \scriptsize~Reference & \scriptsize~Blurry  & \scriptsize~Restormer\cite{zamir2021restormer} & \scriptsize~Restormer-Local
\end{tabular}}
\caption{Qualitative evaluation of our TLC for {Dual-pixel defocus deblurring} on the DPDD dataset~\cite{abuolaim2020defocus}. Restormer with our TLC (\ie, Restormer-Local) more effectively removes blur while preserving the fine image details.}
\label{fig:dualpixel_deblurring}
\end{figure*}

%% file: tabs/Tab_denoise.tex
\begin{table*}[t]
\centering
\caption{
{Gaussian image denoising} comparisons for two kinds of images and two categories of methods on Urban100~\cite{huang2015single} dataset. Top super row: learning a single model to handle various noise levels. Bottom super row: training a separate model for each noise level}
\label{tab:denoise}
\begin{subtable}[t]{0.48\linewidth}
\scriptsize
\centering
\setlength{\tabcolsep}{3pt}
\caption{Gaussian grayscale image denoising} 
\label{table:grayscaledenoising}
\begin{tabular}{l|ccc}
\toprule
{Method}                     & $\sigma$$=$$15$   & $\sigma$$=$$25$   & $\sigma$$=$$50$   \\ \midrule
DnCNN~\cite{zhang2017beyond}                    & 32.28             & 29.80             & 26.35             \\
FFDNet~\cite{zhang2018ffdnet}              & 32.40             & 29.90             & 26.50             \\
IRCNN~\cite{zhang2017learning}        & 32.46             & 29.80             & 26.22             \\
DRUNet~\cite{zhang2021plug}           & 33.44             & 31.11             & 27.96             \\ \midrule
Restormer~\cite{zamir2021restormer}   & {33.67} & {31.39} & {28.33} \\ 
{Restormer-Local}              & \textbf{33.73}             & \textbf{31.48}             & \textbf{28.45}             \\\midrule \midrule
MWCNN~\cite{liu2018multi}             & 33.17             & 30.66             & 27.42             \\
NLRN~\cite{liu2018non}               & 33.45             & 30.94             & 27.49             \\
RNAN~\cite{zhang2019residual}         & -                 & -                 & 27.65             \\
DeamNet~\cite{ren2021adaptive} & 33.37             & 30.85             & 27.53             \\
DAGL~\cite{mou2021dynamic}        & 33.79             & 31.39             & 27.97             \\
SwinIR~\cite{liang2021swinir}         & 33.70              & 31.30              & 27.98             \\ \midrule
Restormer~\cite{zamir2021restormer}   & {33.79} & {31.46} & {28.29} \\
{Restormer-Local}              & \textbf{33.85}             & \textbf{31.55}             & \textbf{28.41}             \\     \bottomrule 
\end{tabular}
    \end{subtable}
\hfill
    \begin{subtable}[t]{0.48\linewidth}
    \scriptsize
    \centering
\caption{Gaussian color image denoising}
\label{table:colordenoising}
\begin{tabular}{l|ccc}
\toprule
{Method}                     & $\sigma$$=$$15$   & $\sigma$$=$$25$   & $\sigma$$=$$50$   \\ \midrule
IRCNN~\cite{zhang2017learning}      & 33.78             & 31.2              & 27.7              \\
FFDNet~\cite{zhang2018ffdnet}            & 33.83             & 31.4              & 28.05             \\
DnCNN~\cite{zhang2017beyond}                  & 32.98             & 30.81             & 27.59             \\
DRUNet~\cite{zhang2021plug}         & 34.81             & 32.60              & 29.61             \\ \midrule
Restormer~\cite{zamir2021restormer} & {35.06} & {32.91} & {30.02} \\
{Restormer-Local}            & \textbf{35.14}             & \textbf{33.01}             & \textbf{30.16}             \\ \midrule \midrule
RPCNN~\cite{xia2020identifying}           & -                 & 31.81             & 28.62             \\
BRDNet~\cite{tian2020image}        & 34.42             & 31.99             & 28.56             \\
RNAN~\cite{zhang2019residual}       & -                 & -                 & 29.08             \\
RDN~\cite{zhang2020residual}             & -                 & -                 & 29.38             \\
IPT~\cite{chen2021pre}              & -                 & -                 & 29.71             \\
SwinIR~\cite{liang2021swinir}       & 35.13             & 32.90              & 29.82             \\ \midrule
Restormer~\cite{zamir2021restormer} & {35.13} & {32.96} & {30.02} \\ 
{Restormer-Local}            & \textbf{35.21}             & \textbf{33.06}             & \textbf{30.17}       \\ \bottomrule 
\end{tabular}
    \end{subtable}

\end{table*}

%% file: imgs/Fig_denoise.tex
\begin{figure}[!t]
\centering
\scalebox{0.99}{
\begin{tabular}[b]{c@{ } c@{ }  c@{ } c@{ } c@{ }   }
\includegraphics[width=.260\textwidth,valign=t]{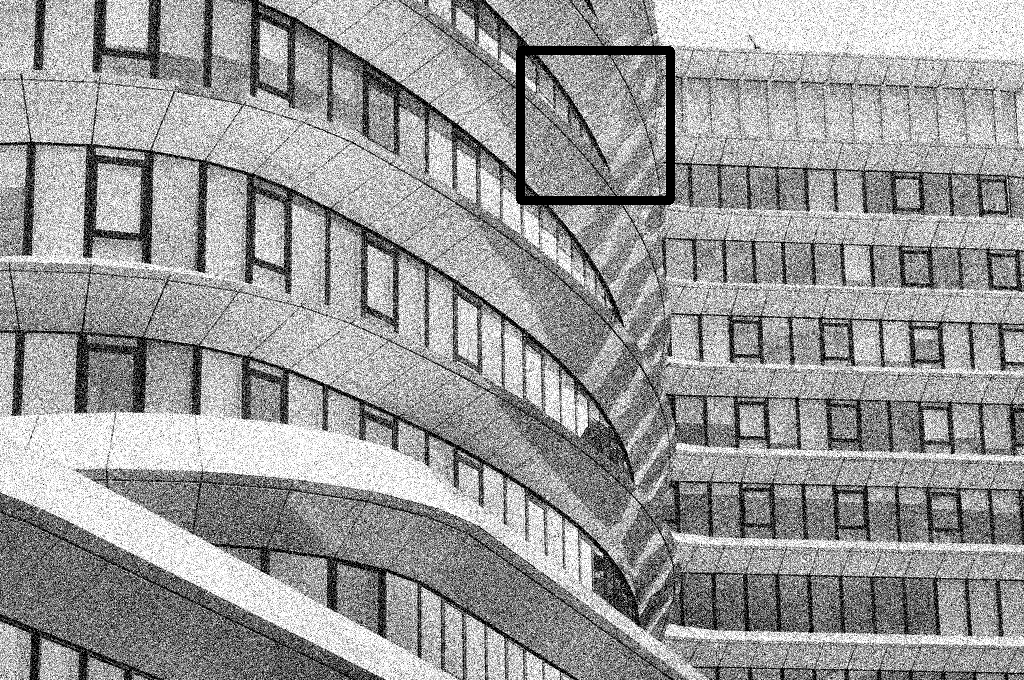}&   
    \includegraphics[width=.173\textwidth,valign=t]{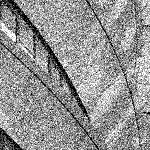} &
    \includegraphics[width=.173\textwidth,valign=t]{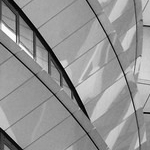} &
    \includegraphics[width=.173\textwidth,valign=t]{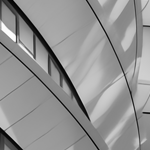} &
    \includegraphics[width=.173\textwidth,valign=t]{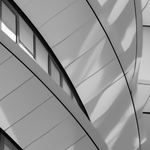}
\\
\scriptsize~14.90 dB   &\scriptsize~14.90 dB &  \scriptsize~PSNR & \scriptsize~31.36 dB & \scriptsize~31.60 dB   \\
   \scriptsize~Noisy Image  & \scriptsize~Noisy & \scriptsize~Reference & \scriptsize~Restormer\cite{zamir2021restormer} & \scriptsize~Restormer-Local  \\

\end{tabular}}
\caption{Qualitative evaluation of our TLC for Gaussian image denoising. Our Restormer-Local removes noise while preserving the fine image details.}
\label{fig:denoise}
\end{figure}

%% file: tabs/Tab_restoration.tex
\begin{table}[t]
\RawFloats
\setlength{\tabcolsep}{3pt}
    \parbox{.48\linewidth}{
    \centering
\caption{Deraining results on SPA-Data~\cite{wang2019spatial} dataset}
 \label{tab:SOTA.derain}
    \begin{tabular}{l|cc}
    \toprule
{Method} & {PSNR$\uparrow$}  & {SSIM$\uparrow$}  \\ \midrule
SPDNet~\cite{fang2020multi}& 43.55 & 0.988   \\
{SPDNet-Local (Ours)} & {43.73} & {0.989} \\ \bottomrule
\end{tabular}
    }
    \hfill
    \parbox{.48\linewidth}{
        \centering
\caption{Dehazing results on outdoor scene images in SOTS~\cite{li2018benchmarking} dataset
}
 \label{tab:SOTA.dehaze}
    \begin{tabular}{l|cc}
     \toprule
{Method} & {PSNR$\uparrow$}            & {SSIM$\uparrow$} \\\midrule
FFANet~\cite{qin2020ffa} & 33.57           & 0.984                 \\
{FFANet-Local (Ours)}      & {33.99}           & {0.985}        \\ \bottomrule 
\end{tabular}
}
\end{table}

%% file: imgs/Fig_derain.tex
\begin{figure}[!t]
\centering
\scalebox{0.99}{
\begin{tabular}[b]{c@{ } c@{ }  c@{ } c@{ } c@{ }   }
\includegraphics[width=.19\textwidth,valign=t]{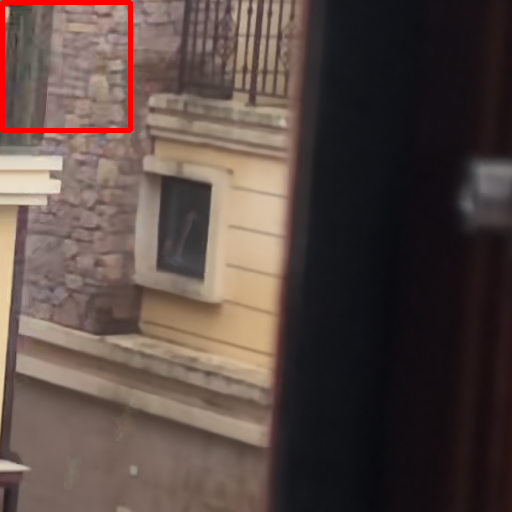}&   
    \includegraphics[width=.19\textwidth,valign=t]{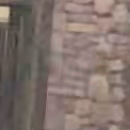} &
    \includegraphics[width=.19\textwidth,valign=t]{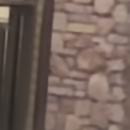} &
    \includegraphics[width=.19\textwidth,valign=t]{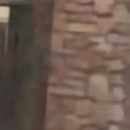} &
    \includegraphics[width=.19\textwidth,valign=t]{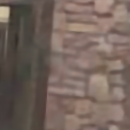}
\\
\small~31.90 dB   &\small~31.90 dB &  \small~PSNR & \small~34.15 dB & \small~38.47 dB   \\
   \small~Rainy Image  & \small~Rainy & \small~Reference & \small~SPDNet~\cite{fang2020multi} & \small~SPDNet-Local  \\

\end{tabular}}
\caption{Qualitative evaluation of our TLC for {image deraining}. SPDNet with our TLC (\ie, SPDNet-Local) superior in the realistic performance of image details and color fidelity.}
\label{fig:deraining}
\end{figure}

%% file: imgs/Fig_dehazing.tex
\begin{figure}[!t]
\centering
\begin{tabular}[b]{c@{ } c@{ }  c@{ } }
\includegraphics[width=.32\textwidth,valign=t]{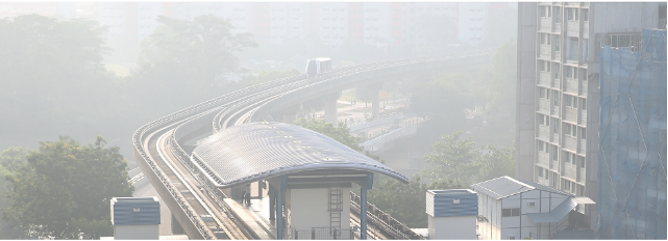}&   
\includegraphics[width=.32\textwidth,valign=t]{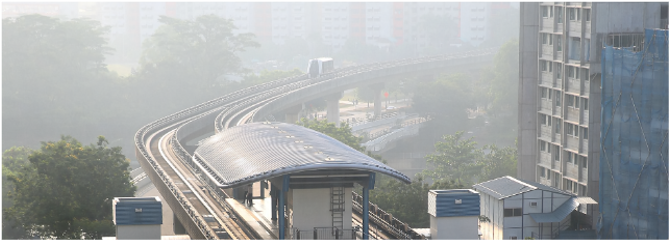}&
\includegraphics[width=.32\textwidth,valign=t]{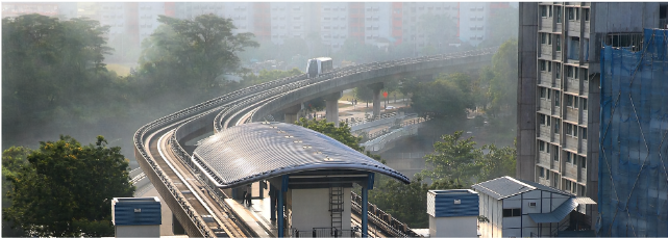}
\\
Hazy Image  & FFANet\cite{qin2020ffa} & FFANet-Local (Ours)
\end{tabular}
\caption{Qualitative evaluation of our TLC for {image dehazing} on realistic hazy image. FFANet with our TLC (\ie, FFANet-Local) generate cleaner result.}
\label{fig:dehazing}
\end{figure}

%% file: tabs/Fig_local.tex
\begin{table}[t]
\centering
\caption{
The results of applying TLC to different modules on GoPro dataset. TLC improves the performance of all models with negligible costs
}
\resizebox{\linewidth}{!}{
\setlength{\tabcolsep}{3.2pt}
\begin{tabular}{c|cc|cc|cc|cc|cc}
\toprule
\multirow{2}{*}{Module} & \multicolumn{2}{c|}{IN~\cite{ulyanov2016instance}} & \multicolumn{2}{c|}{GN~\cite{wu2018group}} & \multicolumn{2}{c|}{GE-$\theta^-$~\cite{hu2018gather}} & \multicolumn{2}{c|}{SE~\cite{hu2018squeeze}}   & \multicolumn{2}{c}{CBAM~\cite{woo2018cbam}}        \\
                        & PSNR$\uparrow$      & MACs$\downarrow$      & PSNR$\uparrow$            & MACs$\downarrow$           & PSNR$\uparrow$            & MACs$\downarrow$           & PSNR$\uparrow$  & MACs$\downarrow$  & PSNR$\uparrow$  & MACs$\downarrow$  \\ \midrule
Origin                  & 30.95     & 62.13G     & 30.91           & 62.13G          & 30.74           & 62.14G          & 30.82 & 62.14G & 30.53 & 62.19G \\
+TLC                   & 31.11     & 62.13G     & 31.03           & 62.13G          & 30.91           & 62.14G          & 31.21 & 62.19G & 31.05 & 62.27G \\ \midrule
$\Delta$                & +0.16     & +0.00G      & +0.12           & +0.00G           & +0.17           & +0.00G           & +0.39 & +0.05G  & +0.52 & +0.08G \\
\bottomrule
\end{tabular}}
\label{tab:Unet_local}
\end{table}

%% file: chaps/conclusion.tex
\section{Conclusion}
In this work, we reveal the global information distribution shifts between training and inference due to train-test inconsistency of global operation, which negatively impacts the performance of restoration model. We propose simple yet test-time solutions, dubbed Test-time Local Converter, which replaces the information aggregation region from the entire spatial dimension to the local window to mitigate the inconsistency between training and inference. Our approach does not require any retraining or finetuning, and boosts the performance of models on various tasks.

%% file: chaps/acknowledgements.tex
~\\

\noindent\textbf{Acknowledgements:} This research was supported by National Key R\&D Program of China (No. 2017YFA0700800) and Beijing Academy of Artificial Intelligence (BAAI).

%% file: chaps/supplementary.tex
\appendix

\noindent\large\textbf{Supplementary Material}
~\\

\setcounter{table}{0}
\renewcommand{\thetable}{A\arabic{table}}
\setcounter{figure}{0}
\renewcommand{\thefigure}{A\arabic{figure}}

In this document, we provide details of comparison between inference with (overlapping) patches and our TLC (Section~\ref{sec:patches_vs_tlc}) and additional visualized results (Section~\ref{sec:visulized}) of our approach and existing methods.

\section{Inference with (overlapping) patches vs our TLC}
\label{sec:patches_vs_tlc}

\begin{table}[b]
\setlength{\tabcolsep}{6pt}
\centering
\vspace{-2mm}
\caption{PSNR of MPRNet on GoPro when test with different methods. Inference time per image is measured on RTX 2080Ti.}
\vspace{-2mm}
\label{tab:tlc_vs_overlap}
\begin{tabular}{ccccc}
\toprule
Input                 & \#Overlap & TLC & PSNR (dB)  & Time (s) \\ \midrule
\multirow{3}{*}{Whole Image}
 & -       & \xmarkg   & 32.66 & 1.60     \\
 & -       & \cmark   & \textbf{33.31} & 2.03     \\ 
    & -       & ~~\cmark$^\dag$   & \textbf{33.31} & \textbf{1.69}     \\ \midrule
\multirow{2}{*}{Overlapping Patches}          & 16      & -   & 33.09 & 2.60     \\
 & 128     & -   & 33.15 & 6.50   \\
\bottomrule
\multicolumn{5}{l}{$^\dag$ denotes optimised implementation of TLC}
\vspace{-3mm}
\end{tabular}
\end{table}

In this section, we compare our TLC with patch inference.
First, Table~\ref{tab:tlc_vs_overlap} experiments on GoPro dataset found that MPRNet model with our test-time method (TLC) achieves higher PSNR (33.31 dB \textit{vs.} 33.15 dB) with less inference time (1.69s \textit{vs.} 6.50s) than inference on overlapping patches cropped from image.
Second, boundary artifacts are also found in the predictions based on overlapping patches (Figure~\ref{fig:artifacts_p256o128}) while our TLC generates natural result without visible artifacts (Figure 2d).
We will describe the details next.

\subsection{Discussion of inference on overlapping patches.} 
Inference with overlapping patches reduces the train-test statistic inconsistency so that it also improves the performance of models. However, it has three main drawbacks:

First, it introduces additional computational costs, as the overlapping regions are restored twice or more by the entire model. While models with our TLC directly restore whole images and TLC has low extra computing costs (Table~7). 

\begin{figure}[t]
\centering
\includegraphics[width=\linewidth]{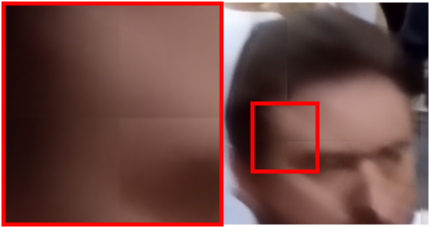}
\caption{MPRNet's deblurring result when inference with overlapping $256\times256$ patches, which also presents vertical and horizontal artifact split lines. Overlapping of different patches is 128.}
\label{fig:artifacts_p256o128}
\vspace{-2mm}
\end{figure}

Second, it can \textit{not} alleviate boundary artifacts on deblurring tasks. We speculate that this is because the global statistics of the two overlapping patches may differ significantly. Restoration of images affected by severe blur requires large receptive field information and the inference of many models (\eg, MPRNet) highly depends on global information.
As a result, predictions of different patches have different estimate of motion blur so that their fusion also show unnatural dividing lines (Figure~\ref{fig:artifacts_p256o128}). 

Third, limited size of patch limits the receptive field of information, which harms the model performance. Pixels in an overlapping region have more range of other pixels for interactions, so that the results based on more overlapping regions are better than fewer ones (Table~\ref{tab:tlc_vs_overlap}). While our TLC utilises full image information and achieves the best results.

\vspace{-5mm}

\subsection{Inference Speed}
Table~\ref{tab:tlc_vs_overlap} shows the inference time of MPRNet with different test methods on RTX2080Ti GPU for a $720\times1280$ image.
With naive implementation using the cumulative sum function provided by Pytorch, TLC introduces 27\% extra times (2.03s \textit{vs.} 1.6s). Note that MPRNet with our TLC still faster and better than inference with overlapping patches. This shows the efficiency of our method.

The cumulative sum operator is unfriendly for GPU because it is a sequential algorithm. As a result, the extra test time caused by TLC will be greater than the theoretical value. A simple way to speed up is to reduce the number of cumulative sum calculations by sampling. Specifically, we can reduce the size of the matrix to $r^2$ times its original size by grid sampling with stride $r$ and use the mean of the sampled matrix to approximate the mean of the original matrix. As a result, the number of calculations needed to do the (one-dimensional) cumulative sum is reduce by $r$ times so that TLC will has $4.78 \times$ faster speed. 
With this careful design, faster TLC only introduces 5.6\% extra times (1.69s \textit{vs.} 1.6s). 

\subsection{Boundary artifacts.}   \label{sec:artifacts}
\begin{figure}[t]
\centering
\includegraphics[width=\linewidth]{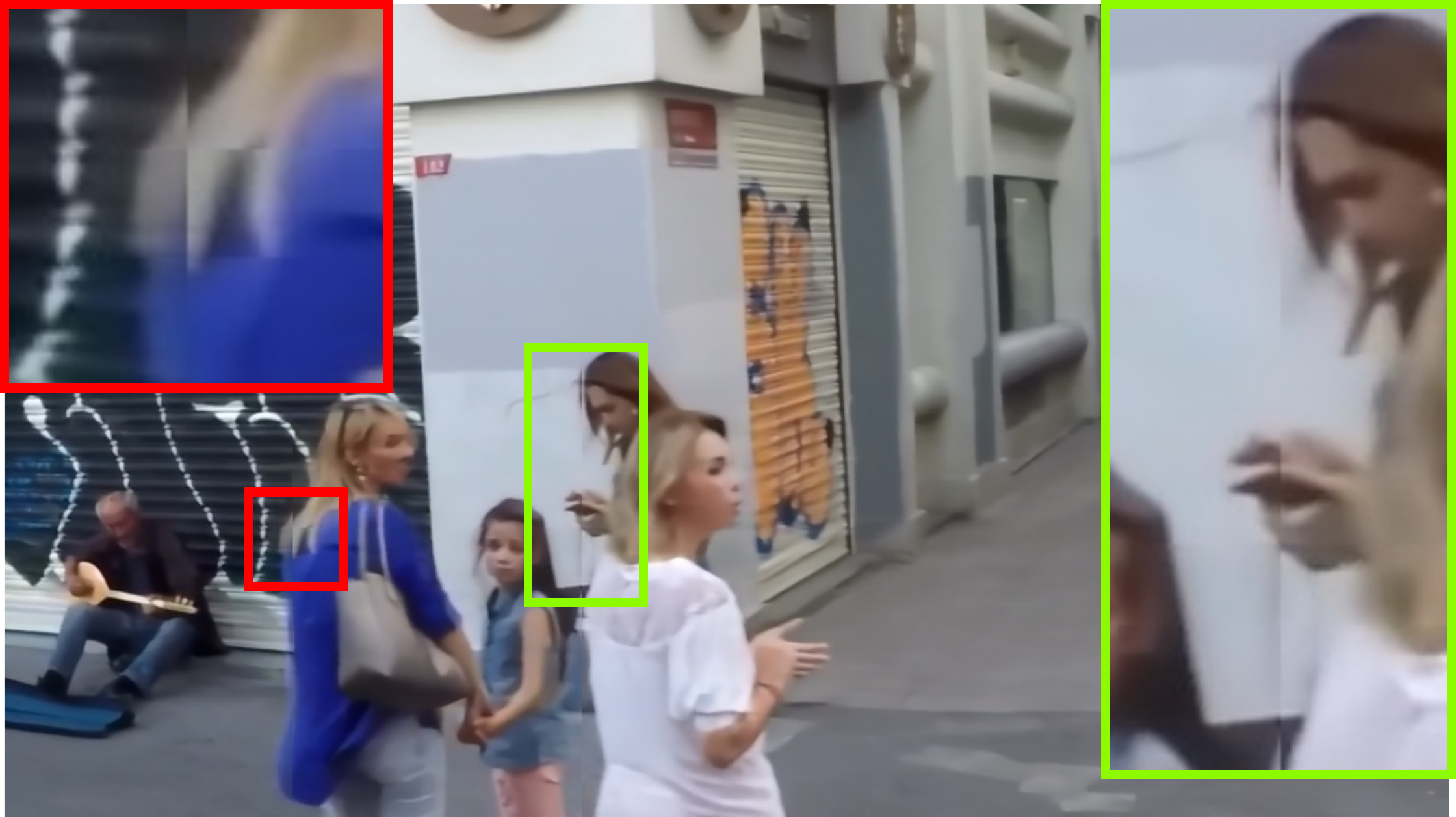}
\caption{
Deblurring results of MPRNet~\cite{zamir2021multi} on GoPro~\cite{nah2017deep}  when inference with patches. Them introduce visible artifacts at patch boundaries, which look like there are vertical stripes cutting through the picture. 
}
\vspace{-2mm}
\label{fig:artifacts}
\end{figure}

\noindent\textbf{Inference with patches}
Cropping the image into patches and predict the result independently induces unsmoothness of the boundary (\ie. ``boundary artifacts'' as demonstrated in~\cite{lee2015block}. We give some example images of block boundary artifacts in Figure~\ref{fig:artifacts}, which are generated by MPRNet~\cite{zamir2021multi} on GoPro~\cite{nah2017deep} dataset. There are obvious boundary artifacts in Figure~\ref{fig:artifacts} which seriously degrades the quality of the image.
 
\noindent\textbf{Inference with overlapping patches.}
We follow the implementation details of SwinIR to do inference with overlapping $256 \times 256$ patches cropped from image and the results of overlapping region are averaged. Boundary artifacts are also found in the deblurred images restored by MPRNet on GoPro dataset. An example is shown in Figure~\ref{fig:artifacts_p256o128} where the blurry image is the same as in Figure~2a (\ie, filename: ``GOPR0384\_11\_05-004042.png''). Compared with our TLC (Figure~2d), inference with overlapping patches (Figure~\ref{fig:artifacts_p256o128}) still introduces visible boundary artifacts.
This phenomenon is different from SwinIR's findings. We attribute this to the nature of severe motion blur and SwinIR did not test on deblurring tasks.

\renewcommand\thesubfigure{\roman{subfigure}}
\begin{figure}[t]
     \centering
     \begin{subfigure}[b]{\textwidth}
         \centering
\includegraphics[width=0.48\linewidth]{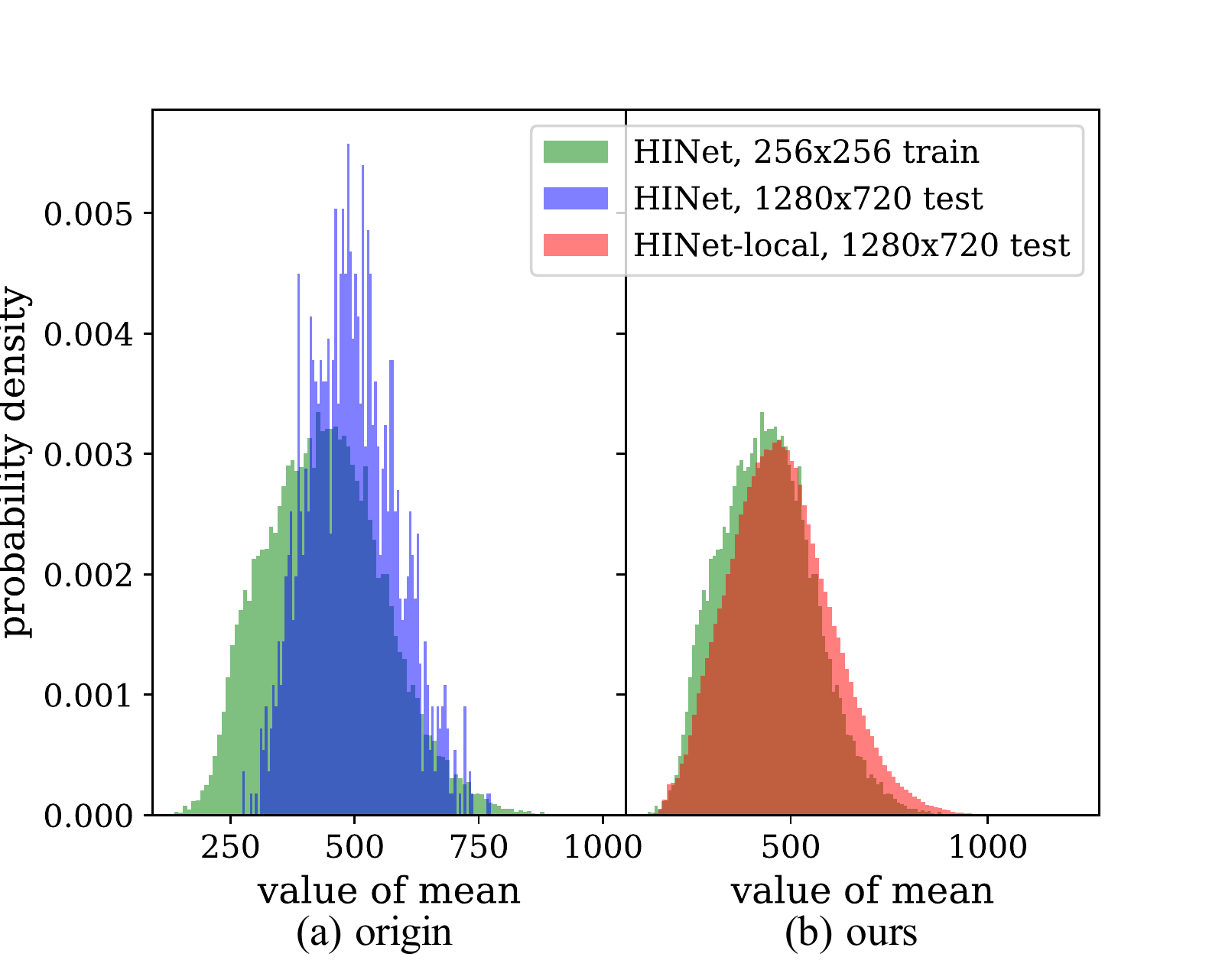}
\includegraphics[width=0.48\linewidth]{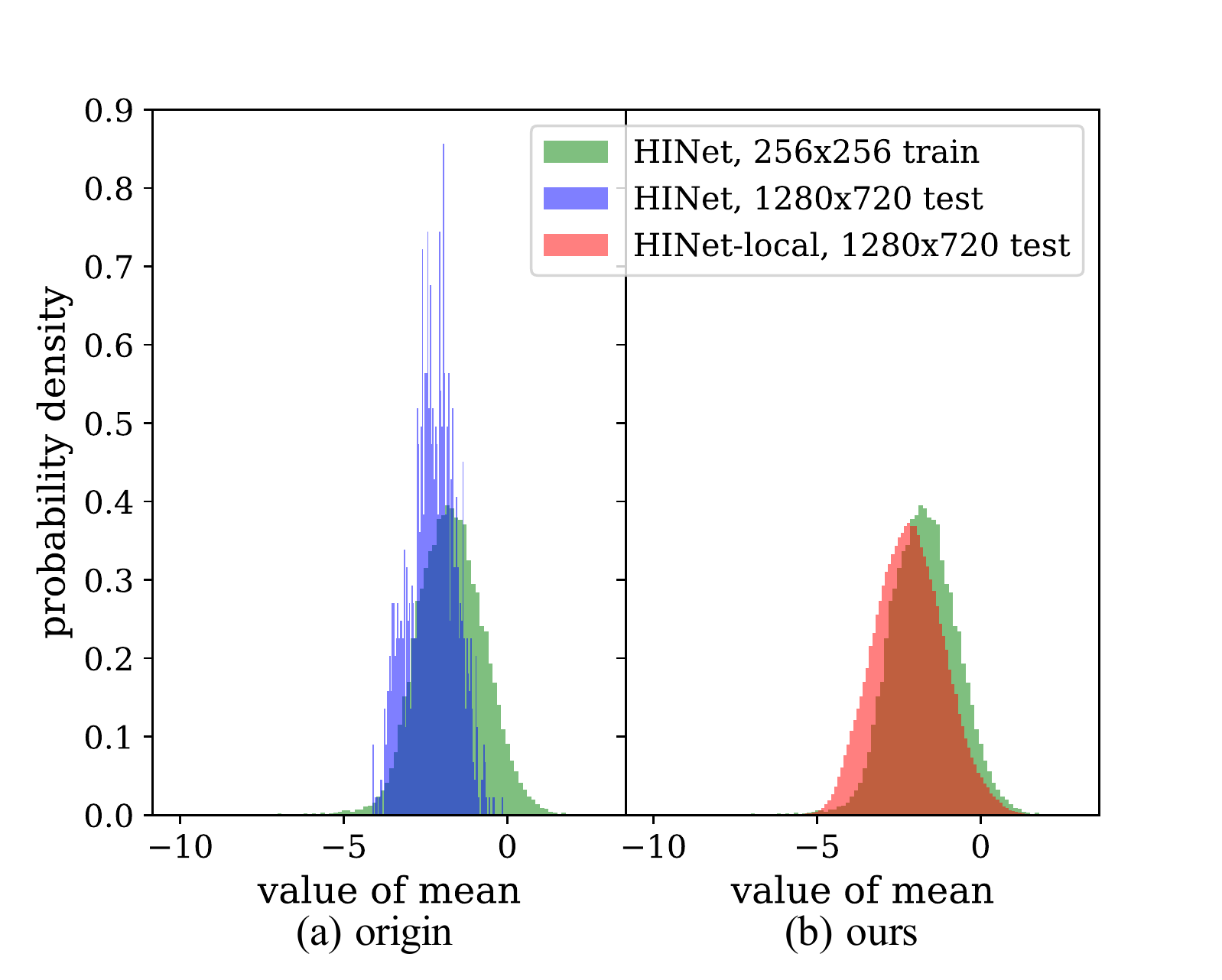}

         \caption{Mean distribution of Instance Normalizations~\cite{ulyanov2016instance} in HINet~\cite{chen2021hinet}}
         \label{fig:HINet_distribution_more}
     \end{subfigure}
     \hfill
     \begin{subfigure}[b]{\textwidth}
         \centering
         \includegraphics[width=0.48\linewidth]{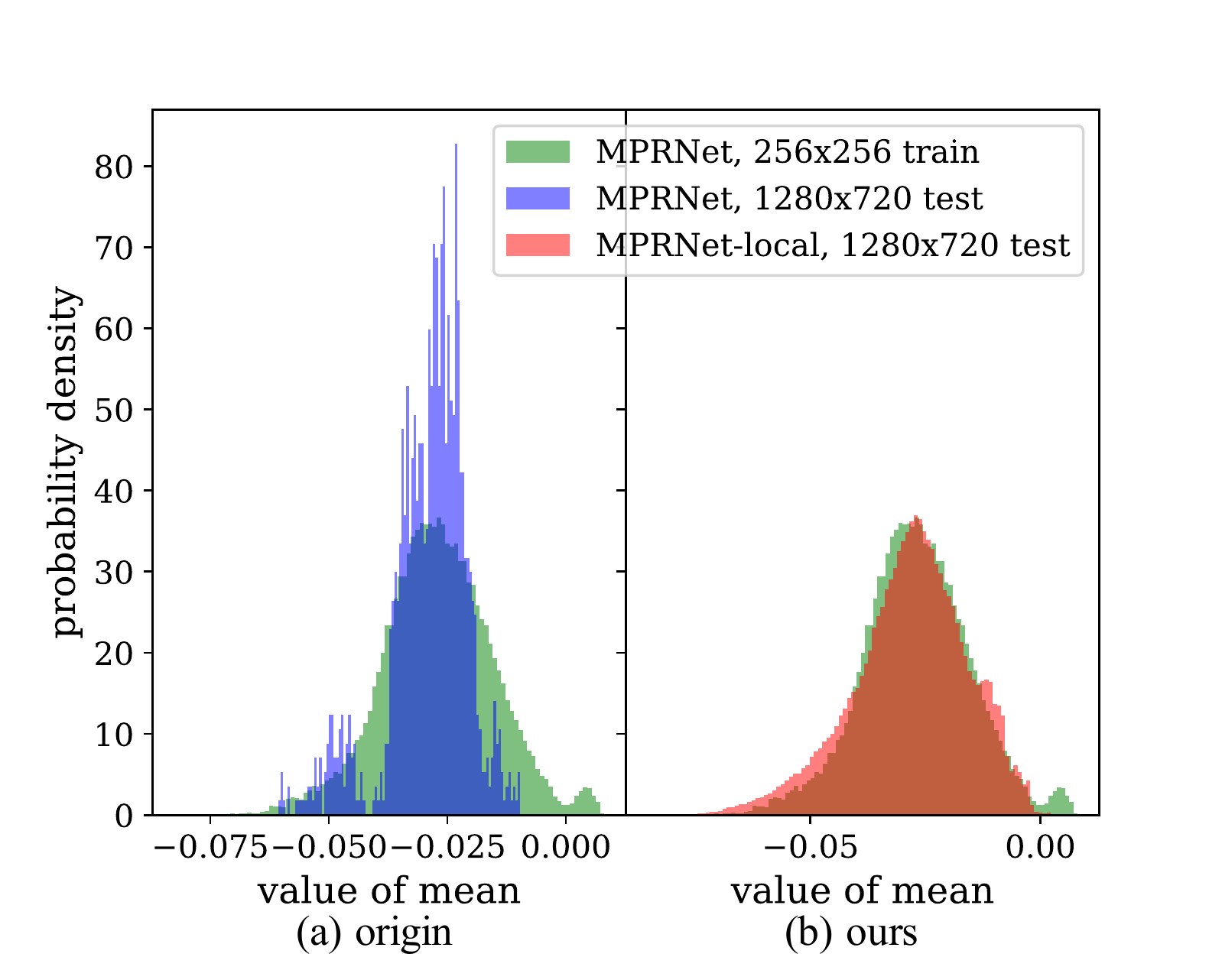}
\includegraphics[width=0.48\linewidth]{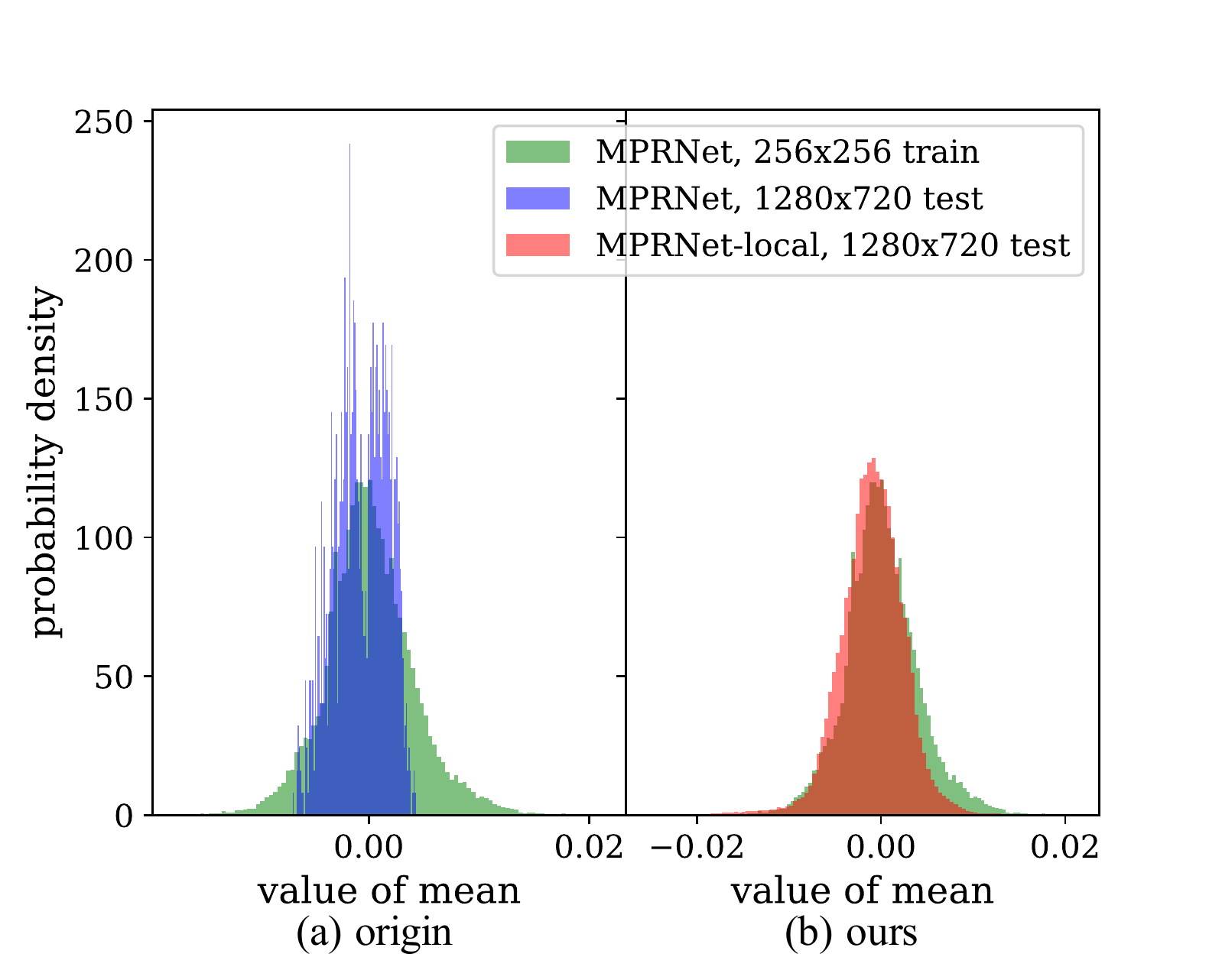}

         \caption{Mean distribution of SE~\cite{hu2018squeeze} layers in MPRNet~\cite{Zamir2021MPRNet}}
         \label{fig:MPRNet_distribution_more}
     \end{subfigure}
        \caption{Visualization of the statistics (mean) distribution of features.
Green: the distribution when training with patches; Blue: the distribution when inference with images; Red: the distribution when inference with images and TLC is adopted (denoted with -local suffix). 
For each sub-figure:  (a) The original test scheme results in distribution shifts. (b) Distribution shifts could be reduced by our proposed TLC.}
        \label{fig:distribution_more}
\end{figure}

\section{Visualized Results} ~\label{sec:visulized}
In this section, we provide additional visual results of statistics distribution (Section~\ref{sec:statistics}) and qualitative comparisons between our approach and existing methods (Section~\ref{sec:qualitative}).

\subsection{Global Information Distribution} \label{sec:statistics}
We provide more results of global information (\ie, mean statistics) distribution between training and inference on GoPro~\cite{nah2017deep} dataset. 
Statistics aggregated by HINet~\cite{chen2021hinet} are shown in Figure~\ref{fig:HINet_distribution_more}. 
Statistics aggregated by MPRNet~\cite{Zamir2021MPRNet} are shown in Figure~\ref{fig:MPRNet_distribution_more}. 
For both HINet~\cite{chen2021hinet} and MPRNet~\cite{Zamir2021MPRNet}, the statistics distribution shifts from training (green) to inference (blue). 
The statistics distribution shifts is reduced by TLC as shown in Figure~\ref{fig:HINet_distribution_more}-b and Figure~\ref{fig:MPRNet_distribution_more}-b compares to Figure~\ref{fig:HINet_distribution_more}-a and Figure~\ref{fig:MPRNet_distribution_more}-a: the statistics distribution obtained by our HINet-local/MPRNet-local (red) is close to the original HINet/MPRNet in the training phase (green).

\subsection{Qualitative Comparisons} \label{sec:qualitative}
In this section, provide additional qualitative results on various image restoration tasks (\eg deburring, deraining and dehazing) for qualitative comparisons.

\begin{figure*}[]
\centering
\includegraphics[width=\linewidth]{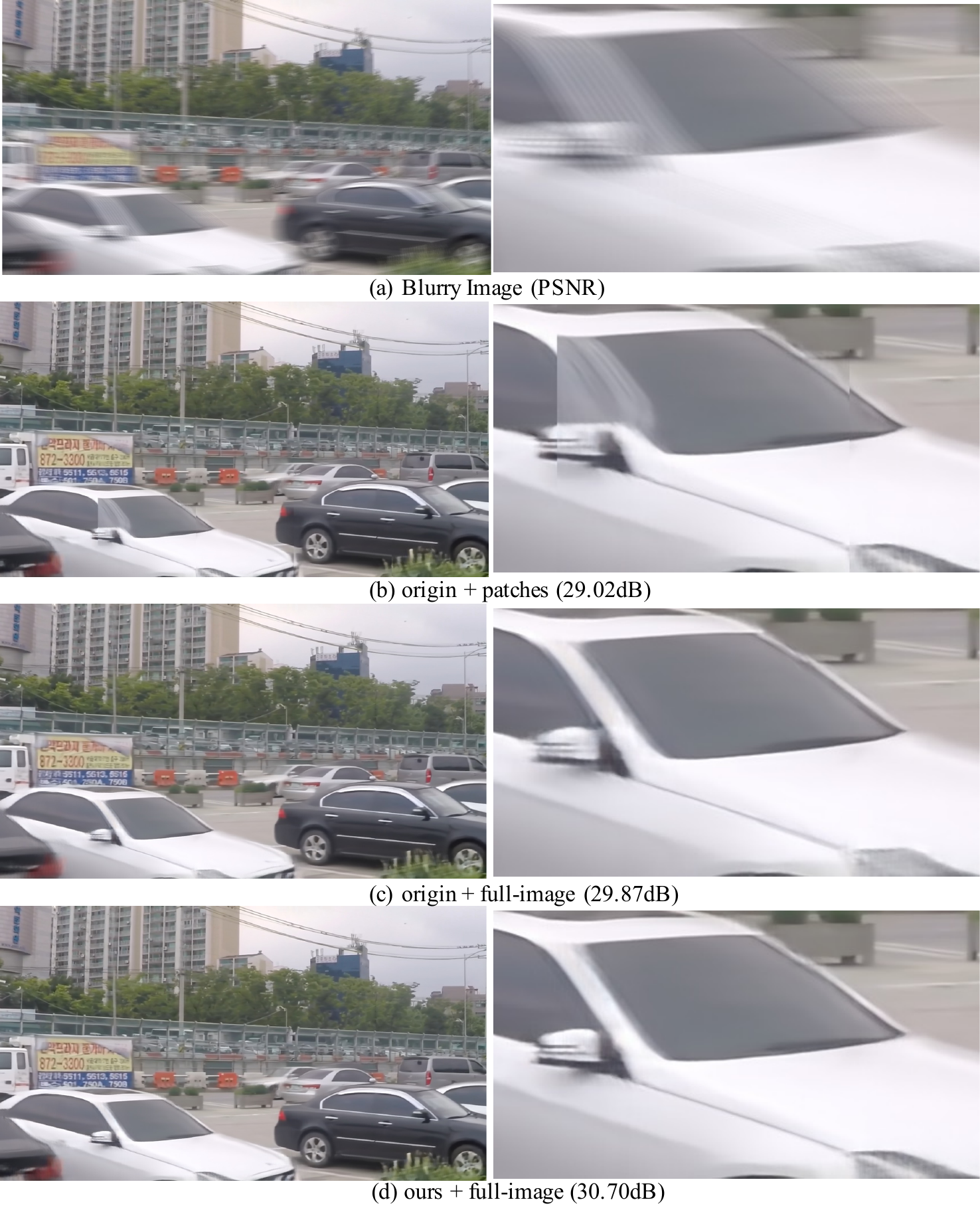}
\caption{
Deblurring results of MPRNet~\cite{zamir2021multi} on GoPro~\cite{nah2017deep} generated by different inference schemes. 
\textbf{Left}: full-images. \textbf{Right}: crops from left image.
(b) Test with patches;
(c) Test with images;
(d) Test with images and TLC is adopted ({ours}). 
It illustrates that (d) provides sharper results than (c) while avoids the boundary artifacts in (b).
}
\label{fig:deblurring_more1}
\end{figure*}

\noindent\textbf{Deburring.}
We give the comparison of the visual effect in Figure~\ref{fig:deblurring_more1} for qualitative comparisons. 
Compared to the original MPRNet~\cite{zamir2021multi} which test with patches (Figure~\ref{fig:deblurring_more1}b), our approach (Figure~\ref{fig:deblurring_more1}d) restores high quality images without boundary artifacts.
Compared to the original MPRNet~\cite{zamir2021multi} which test with images (Figure~\ref{fig:deblurring_more1}c), our approach restores clearer and sharper images.

\begin{figure}[t]
\centering
\includegraphics[width=\linewidth]{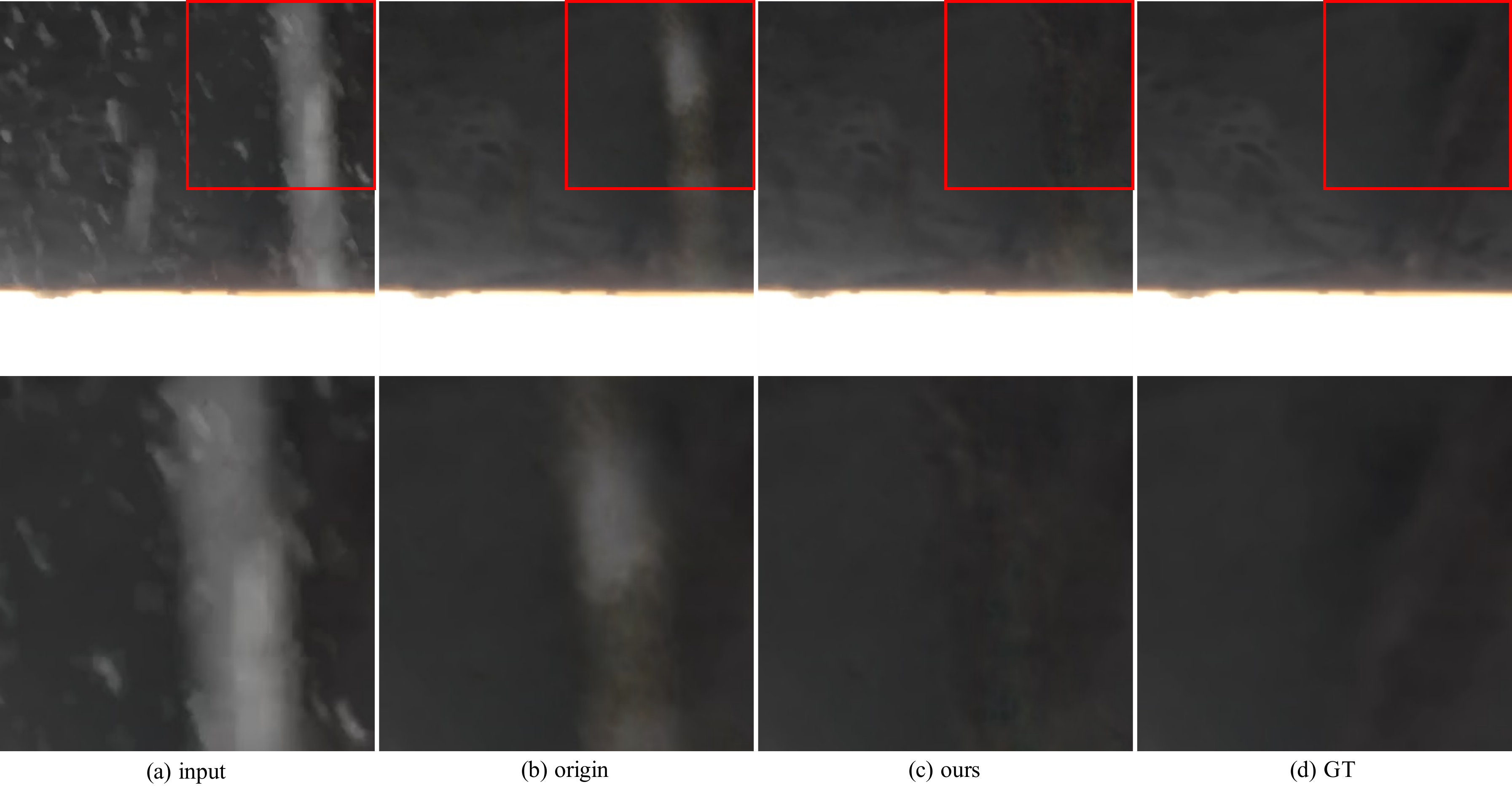}
\caption{
Deraining results of SPDNet~\cite{fang2020multi} on SPA-Data~\cite{wang2019spatial} dataset generated by different inference methods.
(a) Rainy images as inputs. (b) Results based on full-image produced by the original SPDNet.
Some of the rainwater in images is not removed cleanly. (c) Results based on full-image produced by SPDNet with the proposed TLC, which are clearer. (d) Ground truth for reference.
}
\label{fig:deraining_more}
\end{figure}

\noindent\textbf{Deraining.}
We give the comparison of the visual effect in Figure~\ref{fig:deraining_more} for qualitative comparisons. Compared to the original SPDNet~\cite{zamir2021multi}, our approach restores clearer images.

\begin{figure}[t]
\centering
\includegraphics[width=\linewidth]{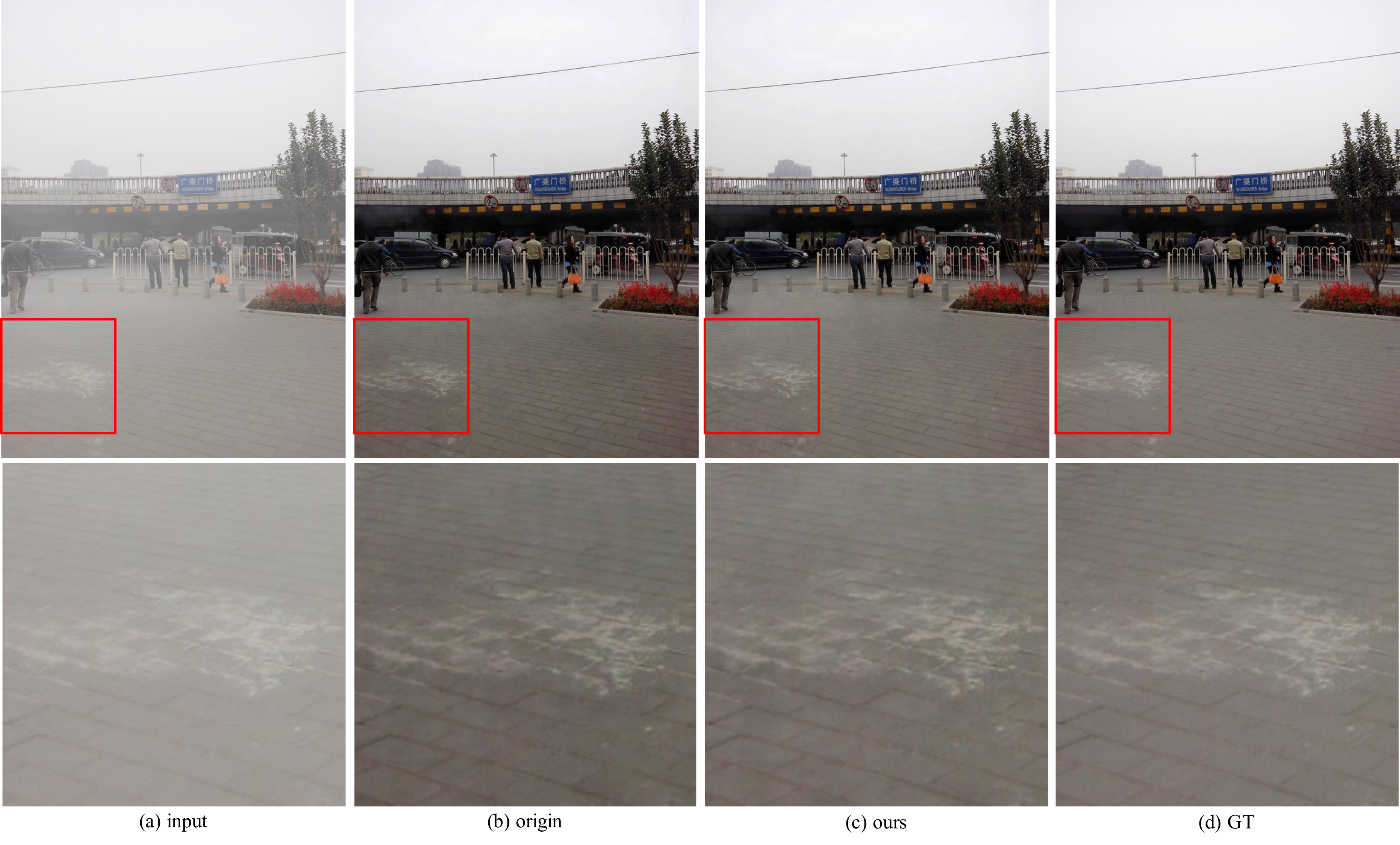}
\vspace{-2mm}
\caption{
Dehazing results of FFANet~\cite{qin2020ffa} on Synthetic Objective Testing Set (SOTS) from RESIDE~\cite{li2018benchmarking} dataset generated by different inference methods.
(a) Hazy images as inputs. (b) Result produced by original FFANet, which is gray with obvious noise. (c) Result produced by FFANet with our TLC, which is brighter with fewer noises. (d) Ground truth for reference.
}
\label{fig:dehazing_more}
\end{figure}

\noindent\textbf{Dehazing.}
We give the comparison of the visual effect in Figure~\ref{fig:dehazing_more} for qualitative comparisons.
Compared to the original FFANet~\cite{zamir2021multi}, our approach restores clearer images.

